\documentclass[lettersize,journal]{IEEEtran}
\usepackage{amsmath,amsfonts}
\usepackage{algorithmic}
\usepackage{algorithm}
\usepackage{array}
\usepackage[caption=false,font=normalsize,labelfont=sf,textfont=sf]{subfig}
\usepackage{textcomp}
\usepackage{stfloats}
\usepackage{url}
\usepackage{verbatim}
\usepackage{graphicx}
\usepackage{tabularx}
\usepackage{cite}
\usepackage{listings}
\usepackage{courier}

\usepackage{caption} % DO NOT CHANGE THIS AND DO NOT ADD ANY OPTIONS TO IT

\usepackage{algorithm}
\usepackage{algorithmic}
\usepackage{amsmath}
\usepackage{amsfonts}
\usepackage{multirow}
\usepackage{pifont}
\usepackage{booktabs}

\usepackage[most]{tcolorbox}

\begin{document}

\title{DEVAL: A Framework for Evaluating and Improving the Derivation
 Capability of Large Language Models}

 \author{
Yifan Li$^{1}$,
Qin Li$^{1}$\thanks{Correspondence to: Qin Li $<$qli@sei.ecnu.edu.cn$>$.},
Min Zhang$^{1}$,
Min Zhang$^{1}$,
Peixin Wang$^{1}$\\
$^{1}$East China Normal University, Shanghai, China
}



\maketitle

\begin{abstract}
Assessing the reasoning ability of Large Language Models (LLMs) over data remains an open and pressing research question. Compared with LLMs, human reasoning can derive corresponding modifications to the output based on certain kinds of changes to the input. This reasoning pattern, which relies on abstract rules that govern relationships between \textit{changes of data}, has not been comprehensively described or evaluated in LLMs. In this paper, we formally define this reasoning pattern as the Derivation Relation (DR) and introduce the concept of Derivation Capability (DC), i.e. applying DR by making the corresponding modification to the output whenever the input takes certain changes. To assess DC, a systematically constructed evaluation framework named \textbf{DEVAL} is proposed and used to evaluate five popular LLMs and one Large Reasoning Model in seven mainstream tasks. The evaluation results show that mainstream LLMs, such as GPT-4o and Claude3.5, exhibit moderate DR recognition capabilities but reveal significant drop-offs on applying DR effectively in problem-solving scenarios. To improve this, we propose a novel prompt engineering approach called Derivation Prompting (DP). It achieves an average improvement of 15.2\% in DC for all tested LLMs, outperforming commonly used prompt engineering techniques.
\end{abstract}


\section{Introduction}
\begin{figure}
  \centering  
  \includegraphics[width=0.46\textwidth]{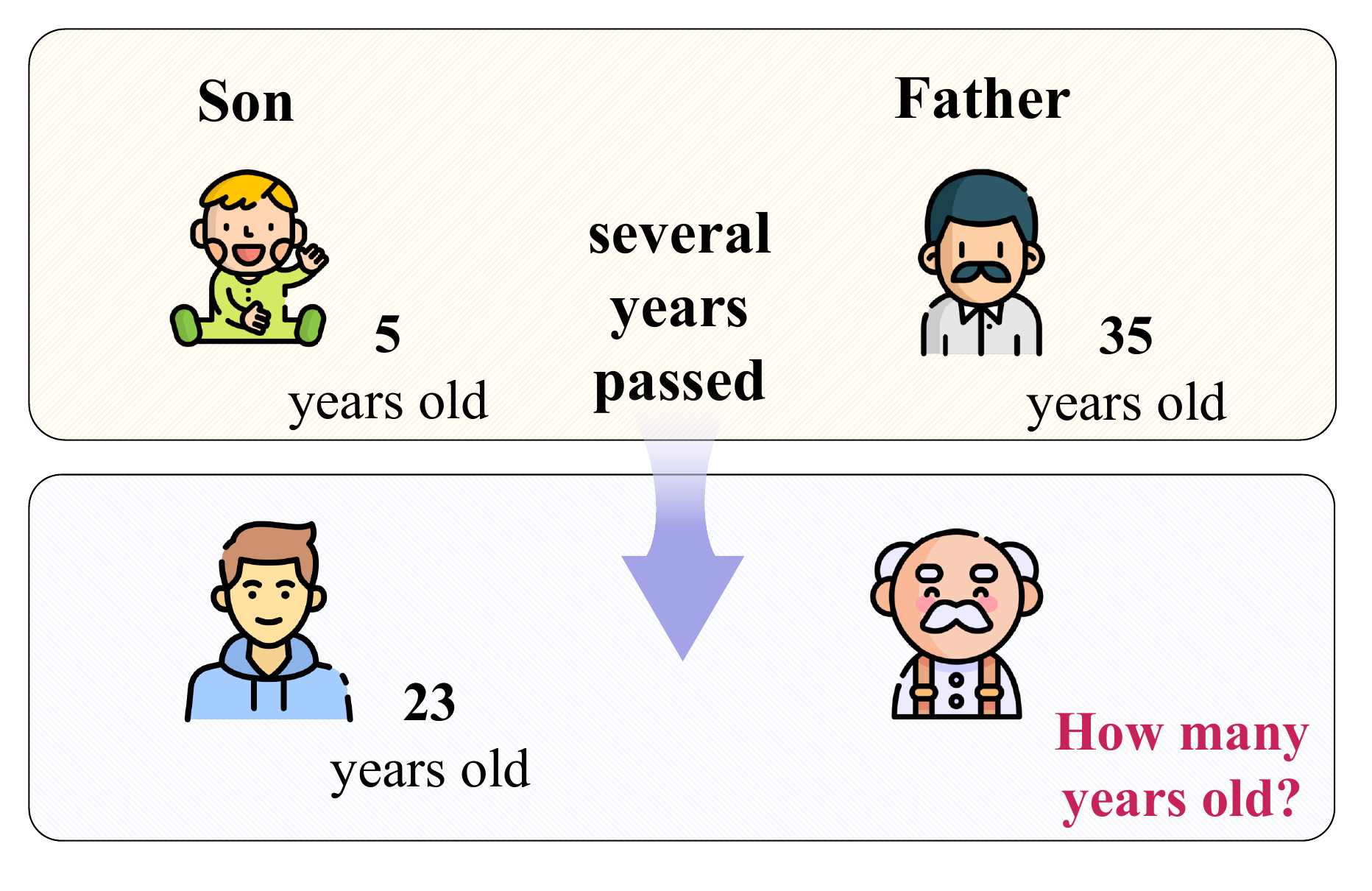}
  \caption{Motivation example of the Derivation Capability.}\label{case}
       
\end{figure}

As Large Language Models (LLMs) evolve into professional agents and begin playing pivotal roles in domain problem solving, evaluating their performance solely on factual correctness has become insufficient.
Instead, abstract reasoning capability has become a more critical aspect of evaluation.
While existing research has explored reasoning strategies such as chain-of-thought~\cite{CoT1} and rule-based inference~\cite{ProntoQA}, these approaches primarily focus on explicit deduction. In addition, human reasoning often relies on recognizing implicit relationships, such as how outputs should systematically change in response to a type of change in inputs. For example, as shown in Fig. \ref{case}, knowing that Mike is 5 and his father is 35 allows us to infer that when Mike is 23, his father should be 53. This demonstrates a crucial pattern of reasoning: abstracting the transformation rule between input and output and applying it to new situations. We refer to this ability as Derivation Capability (DC), i.e., the capacity of LLMs to internalize and apply abstract transformation rules underlying data. Despite its central role in human reasoning, DC remains underexplored in current LLM evaluations.

\begin{figure}
  \centering  
  \includegraphics[width=0.46\textwidth]{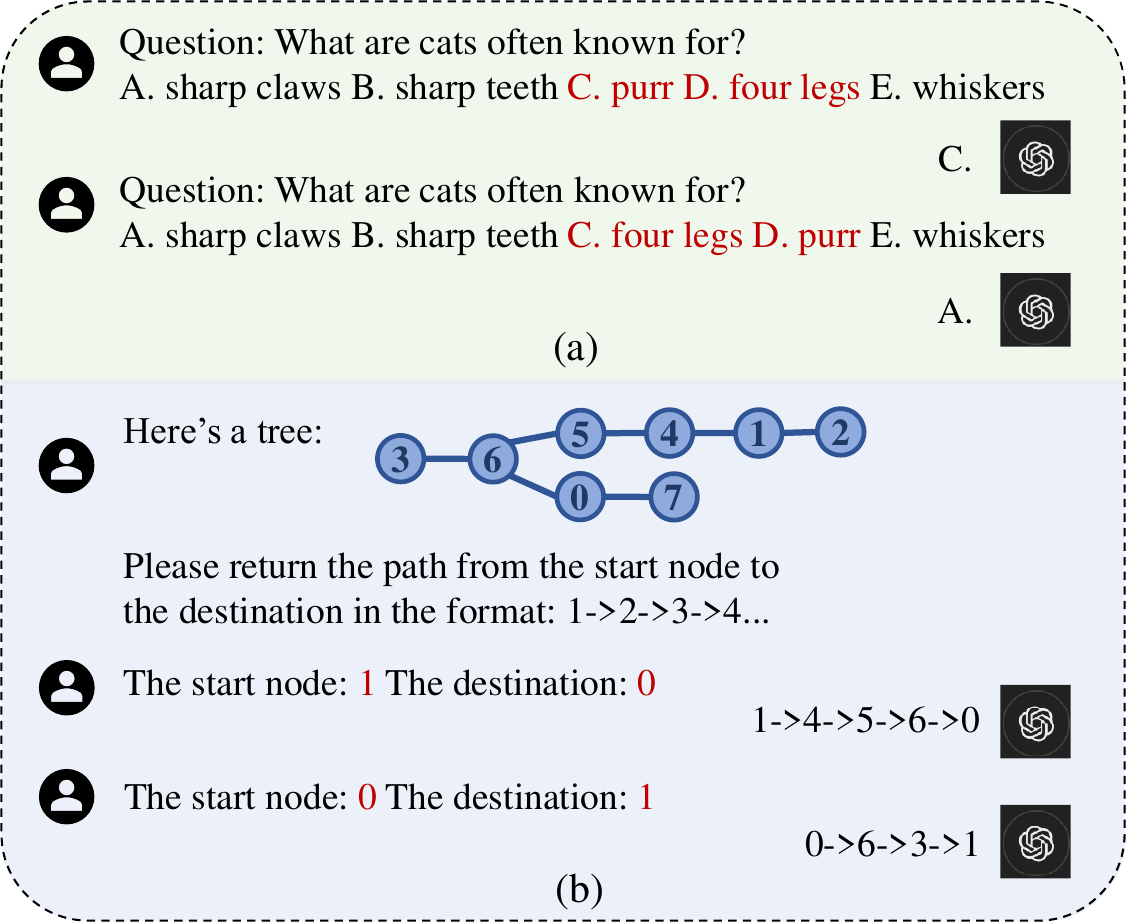}
  \caption{Counterexamples in GPT-4o. We run the test 100 times and take the most frequent answer as the final answer.}\label{motivation}
       
\end{figure}

To investigate DC, we start with simplified tasks where
the transformation rules between inputs and outputs, i.e., the Derivation Relations (DR), are explicitly defined.
In Fig. \ref{motivation}(a), for example, the correct answer should remain the same regardless of the change in options.
However, the LLM incorrectly changes its answer when the options are reassigned.
Similarly, in the path detection task shown in Fig. \ref{motivation}(b), reversing the start and end nodes of a path should yield the same path in reverse. The LLM also fails to adjust its response accordingly.
In these cases, it exhibits an error rate of 62\%, revealing a fundamental weakness: they frequently violate DRs, even in straightforward and well-defined cases. This failure suggests that current LLMs often rely on surface-level cues rather than abstract rule acquisition. By grounding DC in formally defined DRs, we establish a concrete framework, DEVAL, to evaluate whether an LLM truly understands the relational structure underlying tasks.

To enable a rigorous evaluation and improvement of DC, DEVAL provides systematic guidance throughout the evaluation pipeline. Specifically, DEVAL formalizes the DRs for a given task, constructs the corresponding datasets, and defines targeted metrics to quantify DC performance. The core challenge here is that DC is a capability defined over transformations. Hence, traditional accuracy metrics on data do not perform the test. DEVAL addresses this issue by evaluating whether the LLM output behavior obeys the defined rules. This ensures that the evaluation aligns properly with the target property.

We address three central research questions in this paper. \textbf{RQ1:} How can DC be formally defined and evaluated? (Sec. 2) \textbf{RQ2:}
How do LLMs generally perform on DC? (Sec. 3) \textbf{RQ3:} Can DC be improved through prompting strategies? (Sec. 4) We first provide a formal definition of DC and demonstrate how it guides the evaluation.
Then, we construct evaluation datasets covering seven representative tasks, finding that mainstream LLMs fail to apply DR in the responses to related questions.
To mitigate this, we propose a novel Prompt Engineering (PE) method named Derivation Prompting (DP), which explicitly forces DR to be applied and achieve significant improvement.

In summary, the main contributions of this paper are as follows.
\begin{itemize}
    \item We propose a \textbf{formal definition} for the general DR reflecting the link between corresponding changes on inputs and outputs. For a given domain, the general definition can guide us to obtain its concrete instances so that we can use it to evaluate the DC of LLMs.
    \item We propose \textbf{DEVAL}, a framework that generates evaluation datasets based on DR (instead of manual construction through human labeling) in a given task, ensuring that the test data align perfectly with the evaluation properties. The evaluation experiments on a number of common tasks show that the DC of mainstream LLMs is not satisfactory.
    \item We propose a novel prompt engineering approach called \textbf{Derivation Prompting}, which significantly improves the DC of LLMs and outperforms other methods in the evaluated tasks.
\end{itemize}

These findings highlight both the current limitations and the promising pathways to enhance abstract reasoning in LLMs.

\section{Methodology}
In this section, we provide a formal definition of DC, describe the metrics derived from it, and introduce DEVAL, the framework for evaluating DC of LLMs.

\subsection{Definition of Derivation Capability}

% We call the abstract rule that captures the correlations between changes in inputs and outputs the \textbf{Derivation Relation}. It is defined as follows:
% %\subsection*{\S  Definition 1. (Derivation Relation)}
Let $f: X \rightharpoonup Y$ be a partial mapping from set $X$ to set $Y$.
Let \( T \subseteq X \times X \) be a binary relation defined on \( X \), and
 $R \subseteq Y \times Y$ be a binary relation defined on $Y$.
Therefore, the pairs $(X, T)$ and $(Y, R)$ are two relational structures on $X$ and $Y$ respectively, denoted as $\mathcal{X}$ and $\mathcal{Y}$.
We say $T$ and $R$ exhibit the \textit{Derivation Relation} with respect to $f$, denoted as $f \sim (T, R)$, if
and only if $f$ is a homomorphism~\cite{Homo} from $\mathcal{X} \rightharpoonup \mathcal{Y}$, i.e.,
\begin{equation} \small
  \forall x_1,x_2 \in \text{Dom}(f) \cdot (x_1,x_2) \in T \Rightarrow (f(x_1),f(x_2)) \in R
\end{equation} \normalsize
In the definition, $f$ refers to the ground truth of a target problem. It is often approximately represented by a dataset that contains input data and output labels.
$T$ and $R$ refer to changes on the input domain and output domain respectively.
%We use implication from $T$ to $R$ to demonstrate that change $T$ leads to change $R$ causally.
DR infers that when input changes in a particular way, outputs should update correspondingly.
In each different task $f$, we define the corresponding DR, i.e. the pair of $T$ and $R$, as the abstract rule on $f$.

% \begin{figure}[t]
%   \centering
%   \includegraphics[width=0.476\textwidth]{figure2.pdf}
%   \caption{Graphical representation of Derivation Relation and robustness as a special case. In the figure, graph (a) presents the DR, graph (b) presents the robustness, which is the special case of (a).}\label{DR}
% \end{figure}

Based on the definition of DR, we say an LLM $M$ has derivation capability if it can make the right update to the output when the input is changed according to the DR. Note that an LLM behaves in a probabilistic manner: the output is not identical when given the same input. So we use the notation $\hat{M}(x)$ to mean the output from $M$ under input $x$ with the highest probability among the set of all possible outputs, i.e., $\hat{M}(x) = \arg\max_{y} \Pr (M(x) = y)$,
with ties resolved in some way (e.g. by random choice).
% Let $f$ be a partial mapping from $X$ to $Y$, where $f \sim (T, R)$ represents the DR. Let $M$ be an LLM. 
We denote $M_f$ as the LLM under the system prompt which describes the task $f$.
Then we say the LLM $M$ acquires the \textit{derivation capability} on DR $f \sim (T, R)$, denoted as $M \models f \sim (T, R)$, if and only if 
\begin{equation}\small
   \forall x_1,x_2\in \text{Dom}(f)\cdot (x_1, x_2) \in T \Rightarrow (\hat{M}_f(x_1),\hat{M}_f(x_2))\in R
\end{equation}\normalsize

As an illustrative example, we revisit the motivation case presented in Fig. \ref{case}, where T and R are defined as follows:\\[1ex]
\hspace*{5mm}  $T = \{(x_1,x_2)  \mid x_1,x_2 \in \mathcal{N},x_2=x_1 + 18 \}$\\[1ex]
\hspace*{5mm}  $R = \{(y_1,y_2) \mid y_1,y_2 \in \mathcal{N},y_2=y_1 + 18 \}$

\begin{figure*}
  \centering  
  \includegraphics[width=0.99\textwidth]{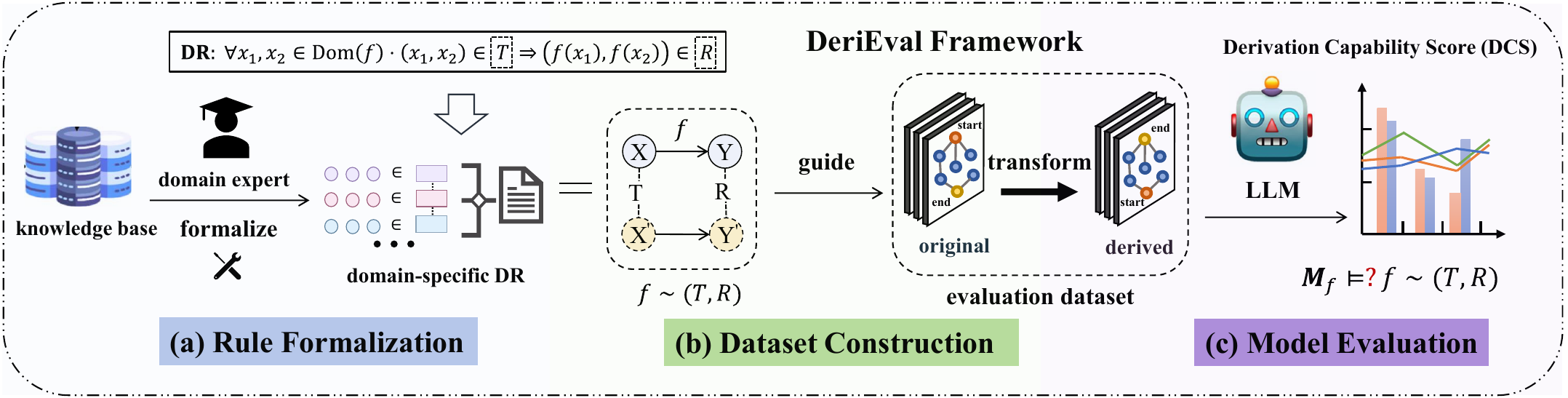}
  \caption{The overall structure of the DEVAL framework.}\label{framework}
   
\end{figure*}

\subsection{Evaluation Metrics}
Due to the infinite number of cases that finite testing cannot cover, it is not possible to prove that LLMs fully satisfy the DC property. Therefore, instead of directly verifying $M \models f \sim (T, R)$
, we assess the extent to which an LLM satisfies the DC requirement from a probabilistic perspective, i.e., demonstrating that $M \mid\!\approx_{\gamma}  f \sim (T, R)$, where $\gamma$ represents the approximation degree, defined as usual as:
\begin{equation}
\gamma = \Pr((\hat{M}_f(x_1),\hat{M}_f(x_2))\in R \mid (x_1,x_2)\in T)
\end{equation}
In practical implementation, the probability can be calculated through sampling.
We use a sampled dataset $D$ with the same distribution to the task inputs, an LLM $M_f$ and DR pair $(T, R)$. With each test case inputs denoted as $x_1 \in D$ and $x_2\in D$, the approximation degree $\gamma$ is then defined as:
\begin{equation}
\gamma = \frac{|\{(x_1,x_2) \mid (x_1,x_2)\in T, (\hat{M}_f(x_1),\hat{M}_f(x_2))\in R\}|}{|\{(x_1,x_2)\mid (x_1,x_2)\in T\}|}
\end{equation}
We refer to the metric $\gamma$ as Derivation Capability Score (DCS), which quantifies the performance in terms of DC.

\subsection{Evaluation Methods}
To evaluate LLM in terms of DC practically, we develop the DEVAL framework to generate
domain-specific definitions and construct the corresponding evaluation datasets. 
As shown in Fig. \ref{framework}, DEVAL consists of three main components: Rule Formalization, Dataset Construction, and Model Evaluation.
The \textbf{Rule Formalization} phase chooses the abstract rule from the task domain. The rule presented in other forms should be formalized in DR, i.e., give a specific definition to the relations $T$ and $R$.
The \textbf{Dataset Construction} phase constructs a dedicated dataset for evaluating DC, ensuring that it correctly conveys the input changes.
A data generation script should be constructed from $T$ and pre-process the original datasets.
The \textbf{Model Evaluation} phase sets the task-specific system prompt and executes each case through two rounds of interaction. An evaluation script should be given by $R$, which quantifies whether the two-round output aligns with the expected relation. Finally, the DCS result is calculated as the proportion of the answers of LLM that pass the evaluation script.

\subsection{Case Study}
In this section, we take Fig. \ref{motivation}(b) as an example to show the whole process in DEVAL. In this path-finding question, let $Tree$ be the set of problem trees with each having nodes $V = \{v_1, v_2, \ldots, v_n\}$ and edges $E \subseteq  V \times V$.
For each case, the LLMs should find the path between $v_1$ and $v_k$, denoted as $\langle v_1,v_2,...,v_k \rangle $. Thus, the input and output domain can be defined as follows:\\[1ex]
\hspace*{5mm}\small  $X = \{(tree,start,end) \mid tree \in Tree ,start, end \in V\} $\\[1ex]
\hspace*{5mm}  $Y = \{\langle v_1,v_2,...,v_k \rangle \mid \forall i \in \{1,2,...,k-1\} \cdot (v_i,v_{i+1}) \in E\}$ \normalsize

In the \textbf{Rule Formalization} phase, DEVAL leverages the symmetry of the problem to observe that swapping the start and end points in the input results in the output path being reversed. This defines the relations $T$ and $R$ as follows:\\[1ex]
\hspace*{5mm} \small  $T = \{(x_1,x_2)  \mid \exists x_1 = (tree,v_1,v_2),  x_2 = (tree,v_2,v_1)\}$\\[1ex]
\hspace*{5mm}   $R = \{(y_1,y_2) \mid y_1 = \langle v_1,...,v_k \rangle, y_2 = \langle v_k,...,v_1 \rangle\}$ \normalsize
\definecolor{orange}{RGB}{68,65,65}

In the \textbf{Dataset Construction} phase, DEVAL collects a sufficient number of initial cases from AQA-Bench~\cite{AQA-Bench}, and constructs a data generation script based on the relation
$T$. This script replaces the start and end points in the original inputs to generate the transformed cases as follows:
\begin{lstlisting}
def data_generation(graph, start, end):
    return (graph, end, start)
\end{lstlisting}

In the \textbf{Model Evaluation} phase, DEVAL constructs an evaluation script based on the relation $R$, which determines whether the two outputs of the LLM are reverses of each other as follows:
\begin{lstlisting}
def evaluation(answer1, answer2):
    return answer1 == answer2[::-1]
\end{lstlisting}

\begin{center}
  \begin{tcolorbox}[colback=gray!10,%gray background
    top=2pt,       
    bottom=2pt,   
    colframe=orange,% black frame colour
    fontupper=\small,
    width=8.5cm,% Use 5cm total width,
    arc=2mm, auto outer arc,
    title={Path Finding},breakable,]
    \definecolor{red}{RGB}{89,38,57}
    \definecolor{green}{RGB}{32,69,85}
    \definecolor{yellow}{RGB}{91,90,37}
  
    \texttt{\textcolor{red}{\textbf{<System Prompt>}}}
    $\mathtt{In\ the\ following\ questions,\ you}$
    $\mathtt{will\ receive\ a\ tree\ a\ start\ point,\ and\ an\ end\ point.}$
    $\mathtt{Please\ return\ the\ path\ from\ the\ start\ to\ the\ end.}$
    \tcbline
    \texttt{\textcolor{yellow}{\textbf{<User Prompt>}}}
    $\mathtt{The\ tree: \{nodes: [0, 1, 2, 3, 4, 5, 6, 7]}$
    $\mathtt{edges: [[0, 2], [1, 3], [2, 6], [2, 4],[3, 5],[3, 6], [4, 7]]\}}$\\
    $\mathtt{The\ start\ node:\ 2\ The\ end\ node:\ 5\ }$
    
    \texttt{\textcolor{green}{\textbf{<Answer>}}}
    % Finally, rounding down the value of e^4 gives us the final answer. Therefore, the final answer is: 54."
    $\mathtt{2\rightarrow 6\rightarrow 3\rightarrow 5}$
    \vspace{-4pt}
    \tcbline
    \vspace{-4pt}
    \texttt{\textcolor{yellow}{\textbf{<User Prompt>}}}
    $\mathtt{The\ tree: \{nodes: [0, 1, 2, 3, 4, 5, 6, 7]}$
    $\mathtt{edges: [[0, 2], [1, 3], [2, 6], [2, 4],[3, 5],[3, 6], [4, 7]]\}}$\\
    $\mathtt{The\ start\ node:\ 5\ The\ end\ node:\ 2\ }$
    
    \texttt{\textcolor{green}{\textbf{<Answer>}}}
    % Finally, rounding down the value of e^4 gives us the final answer. Therefore, the final answer is: 54."
    $\mathtt{5\rightarrow 3\rightarrow 6\rightarrow 2}$
  \end{tcolorbox}
  \end{center}
\setcounter{figure}{4}
\begin{center}
Figure 4: Example of DEVAL applied to the path-finding.
\end{center}

As shown in Fig. 4, the two-round answer passes the evaluation script, and thus this is a positive example. This case study demonstrates the ability of the DEVAL pipeline to systematically formalize DR, generate evaluation data, and evaluate LLM results in a real-world reasoning scenario.

\section{Derivation Capability Evaluation}

In this section, we focus on the performance and error cases of DC. We first provide a comprehensive overview of overall DC performance across all datasets and then perform an attribution analysis based on the reasoning process.

\subsection{DC Performance Evaluation on LLMs}

We use DEVAL to evaluate six LLMs, including GPT-3.5, GPT-4o, Claude3.5, Qwen, Kimi
% (with versions \texttt{gpt-3.5-turbo}, \texttt{gpt-4o-2024-08-06}, \texttt{claude-3-5-sonnet-20240620}, \texttt{qwen-turbo-2024-11-01}, \texttt{moonshot-v1} respectively)
and O1-mini.
% (\texttt{o1-mini-2024-09-12})
We selected seven original datasets from different domains, covering a variety of task types, namely choice, image, code, logic, math, algorithm, and space. For each task, DEVAL balanced different input scales, identified three types of rules, and constructed opposite DRs, namely \textbf{\textit{ID} (IDentity), \textit{GE} (GEneral), and \textit{TS} (Task-Specific).}
Typically, \textit{ID} means the special case in DR that required the output to remain unchanged. \textit{GE}
uses domain-independent rules (e.g., symmetry, equivalence replacement) to construct general-purpose DRs. \textit{TS} applies domain-specific rules (e.g., mathematical theorems, logical rules) to construct task-specific DRs. 
All construction and evaluation procedures follow the methodology described in Sec. 2.3, with detailed implementation provided in Appendix B.

\begin{figure}[t]
\centering
  \includegraphics[width=0.476\textwidth]{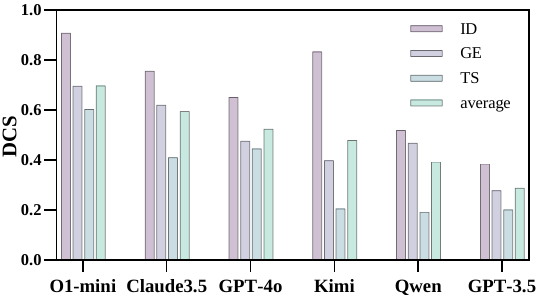}
\caption{DC performance in different LLMs and DR types.}\label{5}
       
\end{figure}

As shown in Fig. \ref{5},
\textbf{most LLMs perform poorly in DC.} Among them,
O1-mini achieves the highest performance with an average score of 69.8\%, followed by Claude3.5
(59.6\%), GPT-4o (52.5\%), Kimi (48.0\%), Qwen (39.4\%) and GPT-3.5 (28.9\%).
Most LLMs exhibit poor performance, with an average score below 60\%.
In addition, the average performance in \textit{GE} and \textit{TS} is 49.1\% and 34.3\%, respectively,
indicating that current LLMs still struggle to maintain consistency between transformations.

% \begin{figure}[t]
%   \centering  
%   \includegraphics[width=0.476\textwidth]{figure7.pdf}
%   \caption{Usage-cost evaluation for LLMs.}\label{RQ2-3}
%        
% \end{figure}

% \textbf{High performance comes at a high cost.} As shown in Fig. \ref{RQ2-3}, while O1-mini achieves the best performance (with an average score of 69.8\%), its usage cost is significantly higher than that of other LLMs. As an OLM, the internal self-feedback iterations of O1-mini contribute to its superior capabilities but also lead to higher token consumption. When considering costs, O1-mini offers little advantage in terms of cost-effectiveness. GPT-4o and Claude3.5, on the other hand, provide a balanced alternative, delivering competitive performance at a lower cost.

\subsection{Error Analysis of DC Evaluation}
To better identify the root causes of failures and lay the groundwork for subsequent performance improvement strategies, we conducted an in-depth analysis of representative failure cases. Specifically, we extracted a set of incorrect responses and prompted LLMs to generate their reasoning chains in the form of CoT. By examining the intermediate reasoning steps, we were able to trace and categorize the sources of errors.

\begin{figure}[t]
\centering
    \includegraphics[width=\linewidth]{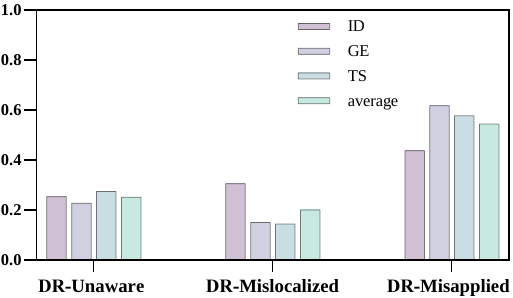}
  \caption{Error pattern evaluation in different DR types.}\label{6}
       
\end{figure}

Our analysis reveals that LLMs tend to exhibit structured and recurring failure types. We classify these errors into three categories. \textbf{DR-Unaware:} LLM keeps the same output. It does not
recognize and incorporate the input transformation defined by $T$. \textbf{DR-Mislocalized:} LLM changes the output but fails to determine how it should change, i.e., the relation $R$. \textbf{DR-Misapplied:} LLM correctly identifies how the output should change, but the final result is wrong.

To operationalize this classification, we employed GPT‑4o as an evaluator to analyze each step of the CoT output and map it to the corresponding error attribution category. We first provided GPT‑4o with the definitions of DC, the error type taxonomy, and representative examples, enabling systematic identification of error types across tasks. We then applied this procedure to all evaluated tasks and computed the distribution of error categories, thereby revealing the prevalent failure modes.

\begin{center}
  \begin{tcolorbox}[colback=gray!10,%gray background
    top=2pt,       % 上边内边距
    bottom=2pt,    % 下边内边距
    colframe=orange,% black frame colour
    fontupper=\small,
    width=8.5cm,% Use 5cm total width,
    arc=2mm, auto outer arc,
    title={Math},breakable,]
    \definecolor{red}{RGB}{89,38,57}
    \definecolor{green}{RGB}{32,69,85}
    \definecolor{yellow}{RGB}{91,90,37}
  
    \texttt{\textcolor{red}{\textbf{<System Prompt>}}}
    %Given the following function expressions, you need to calculate the integral of the function, then substitute [x=4] to the result.
    $\mathtt{Given\ the\ following\ function,}$
    $\mathtt{please\ calculate\ the\ integral\ of\ the\ function,}$
    $\mathtt{then\ substitute\ [x=4]\ to\ the\ result.}$
    \tcbline
    \texttt{\textcolor{yellow}{\textbf{<User Prompt>}}}
    $\mathtt{The\ function\ is:\ e^x.}$
    
    \texttt{\textcolor{green}{\textbf{<Answer>}}}
    % Finally, rounding down the value of e^4 gives us the final answer. Therefore, the final answer is: 54."
    $\mathtt{…\ Finally,\ rounding\ down\ the\ value\ of}$
    $\mathtt{e^4\ gives\ us\ the\ final\ answer.\ Therefore,\ the\ final}$
    $\mathtt{answer\ is:\ 54.}$
    \tcbline
    \texttt{\textcolor{yellow}{\textbf{<User Prompt>}}}
    $\mathtt{The\ function\ is:\ e^x+x.}$
    
    \texttt{\textcolor{green}{\textbf{<Answer>}}}
    % the value of e^4 + 8 gives us the final answer. Therefore, the final answer is: 59.
    $\mathtt{…\ rounding\ down\ the\ value\ of\ e^4 + 8}$
    $\mathtt{gives\ us\ the\ final\ answer.\ Therefore,\ the\ final}$
    $\mathtt{answer\ is:\ 59.}$
  \end{tcolorbox}
  \end{center}

\setcounter{figure}{7}
\begin{center}
Figure 7: Error example of DR-Misapplied case.
\end{center}

As shown in Fig. \ref{6},
  DR-Misapplied emerges as the predominant source of errors. Specifically, DR-Misapplied accounts for approximately 54.57\%, significantly higher than the DR-Unaware and DR-Mislocalized categories, which account for 25.29\% and 20.14\%, respectively.   
  This indicates that \textbf{although LLMs can infer DR to some extent, they often fail to connect these inferred conclusions when responding to subsequent related problems} (as in Fig. 7).
  Moreover, DR-Misapplied remains consistently prevalent across tasks. This stability suggests that DR-Misapplied serves as a promising and generalizable dimension for improvement. In Sec. 4, we aim to reinforce the attention to associations between problems within prompts to prevent disruptions in the reasoning chains between consecutive answers.

\section{Derivation Capability Improvement}
To address the prevalent failure in maintaining DR, we propose a novel prompt engineering (PE) method, Derivation Prompting (DP).
This approach is motivated by the observation in Sec. 3.3 that LLMs often fail to incorporate DR rules into their final answer.
\textbf{DP aims to mitigate this issue by explicitly guiding LLMs through three steps of DR utilization, forcing LLMs to internalize and apply the DR they find when answering two related questions.}
In practice, we taught LLM in natural language descriptions using the following template:
``For the question, you need to (1) explain what change has occurred in the input; (2) explain what kind of change in output should result from this change in input in this task; (3) apply this output change to your original answer."
We compared DP with five other PE methods: Chain-of-Thought (CoT) ~\cite{CoT1}, Step-Back (SB)~\cite{Step-back}, One-Shot (OS), Few-Shot (FS), and ANalogy (AN)~\cite{Analogy} to evaluate its effectiveness in enhancing DC performance.
The specific implementation details are described in Appendix C.

\begin{center}
  \begin{tcolorbox}[colback=gray!10,%gray background
    top=2pt,       % 上边内边距
    bottom=2pt,    % 下边内边距
    colframe=orange,% black frame colour
    fontupper=\small,
    width=8.5cm,% Use 5cm total width,
    arc=2mm, auto outer arc,
    title={Math},breakable,]
    \definecolor{red}{RGB}{89,38,57}
    \definecolor{green}{RGB}{32,69,85}
    \definecolor{yellow}{RGB}{91,90,37}
  
    \texttt{\textcolor{red}{\textbf{<System Prompt>}}}
    %Given the following function expressions, you need to calculate the integral of the function, then substitute [x=4] to the result.
    $\mathtt{Given\ the\ following\ function,}$
    $\mathtt{please\ calculate\ the\ integral\ of\ the\ function,}$
    $\mathtt{then\ substitute\ [x=4]\ to\ the\ result.\ (For\ the}$
    $\mathtt{second\ question,\ you\ need\ to: (1)\ ... \ (2)\ ...\ (3)\ ...)}$
    \tcbline
    \texttt{\textcolor{yellow}{\textbf{<User Prompt>}}}
    $\mathtt{The\ function\ is:\ e^x.}$
    
    \texttt{\textcolor{green}{\textbf{<Answer>}}}
    % Finally, rounding down the value of e^4 gives us the final answer. Therefore, the final answer is: 54."
    $\mathtt{…\ Finally,\ rounding\ down\ the\ value\ of}$
    $\mathtt{e^4\ gives\ us\ the\ final\ answer.\ Therefore,\ the\ final}$
    $\mathtt{answer\ is:\ 54.}$
    \tcbline
    \texttt{\textcolor{yellow}{\textbf{<User Prompt>}}}
    $\mathtt{The\ function\ is:\ e^x+x.}$
    
    \texttt{\textcolor{green}{\textbf{<Answer>}}}
    % the value of e^4 + 8 gives us the final answer. Therefore, the final answer is: 59.
    $\mathtt{(1)\ input\ function\ has\ an\ extra\ term\ [x].}$ \\
    $\mathtt{(2)\ This\ term\ needs\ to\ be\ integrated.}$ \\
    $\mathtt{(3)\ The\ previous\ result\ is\ 54.\ Integrating\ the\ [x]}$ \\
    $\mathtt{term\ gives\ x^2/2.\ Substituting\ x=4\ gives\ 8,\ so}$ \\
    $\mathtt{the\ final\ answer\ is:\ 54+8=62.}$
  \end{tcolorbox}
  \end{center}

\begin{center}
Figure 9: Example of DR-Misapplied case corrected by DP.
\end{center}

\begin{figure}[t]
  \centering  
  \includegraphics[width=0.476\textwidth]{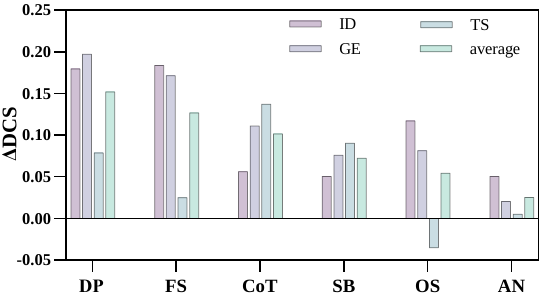}
  \caption{Improvement of DCS under different PE methods.}\label{8}
       
\end{figure}

As shown in Fig. \ref{8},
\textbf{DP shows the most significant improvement.} Specifically, it ranks first in improving the DCS, \textbf{with a gain of 15.2\%},
followed by FS, CoT, SB, OS and AN.
Across all PE methods, the average improvement in DCS is 6.7\%, demonstrating the effectiveness of PE approaches in improving the DC of LLMs.
In addition,
CoT and SB show increasing gains from \textit{ID} to \textit{TS}.
In contrast, reference-based methods such as FS and OS show a downward trend from \textit{ID} to \textit{TS}, indicating that data binding is useful for improving \textit{ID} scenarios but struggles to enhance the awareness of the reasoning structure behind DR.
Meanwhile, DP corrects many errors that CoT previously made, as shown in Fig. 9.
This leads to substantial gains in tasks like math, with improvements of 5.1\%, further underscoring the rationality and effectiveness of our improvement design.
\section{Discussion}
In this section, we discuss the broader implications of our evaluation results and the DEVAL framework.
We first validate the reliability of DEVAL by comparing the results on \textit{ID} type with existing robustness evaluation frameworks, confirming that our conclusions are consistent. This consistency supports the credibility of DEVAL’s findings in the \textit{GE} and \textit{TS} types as well (which are beyond robustness).
Next, we examine whether LLMs can autonomously formulate high‑quality DRs to reduce the amount of manual effort required in the DEVAL framework, and analyze the limitations revealed by our experiments.
Finally, we investigate whether LLMs can acquire DR knowledge through supervised fine‑tuning (SFT) with paired training data, and compare their effectiveness to our DP method, providing insights on further contributions to improving derivation reasoning.
\subsection{DEVAL Framework Validation}

\setcounter{figure}{9}
\begin{figure*}
    \centering  
    \includegraphics[width=1\textwidth]{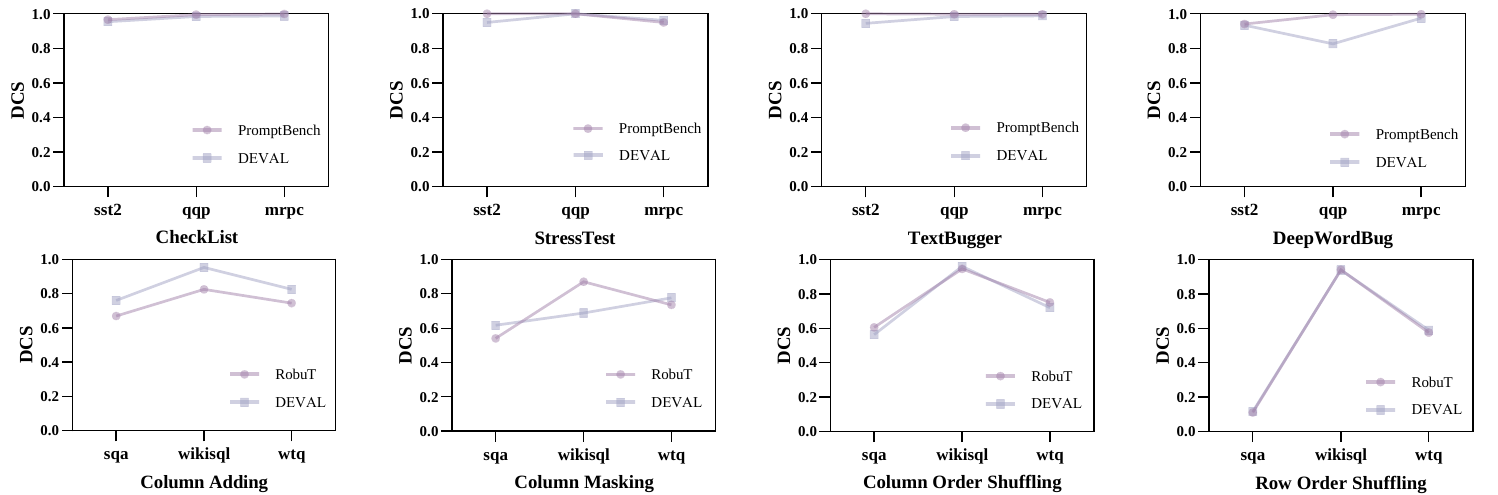}
    \caption{Comparison chart of results in DEVAL and other frameworks.}\label{10}
     
  \end{figure*}

To establish the validity of the DEVAL, we leverage the fact that robustness evaluation can be viewed as a special case of DC when output remains invariant (i.e., the \textit{ID} type). Based on this connection, we select two widely adopted robustness evaluation frameworks (PromptBench and RobuT) as boundary references. By aligning their perturbation settings with the DR formalism, we aim to check whether DEVAL produces consistent and expected results under equivalent conditions. Strong agreement would indicate that DEVAL can reliably capture robustness behavior, thus supporting its broader applicability to general DC evaluation. Task descriptions and DR definitions are provided in Appendix A.

\textbf{PromptBench}~\cite{PromptBench} is a robustness evaluation framework for LLMs against prompt perturbations. In our experiment, we selected four transformation methods (CheckList~\cite{CheckList}, StressTest~\cite{StressTest}, TextBugger~\cite{TextBugger}, and DeepWordBug~\cite{DeepWordBug}) and three datasets (\textit{SST-2}, \textit{QQP}, and \textit{MRPC}) for evaluation. We compared the performance of the same model \texttt{T5-large}~\cite{T5-large} in the robustness evaluation framework and our designed DR. 

\textbf{RobuT}~\cite{RobuT} is another robustness evaluation framework on table-based question answering tasks, where LLMs are required to derive conclusions from table data to answer the question. It evaluates robustness by adding perturbations to table question-answering datasets (WTQ~\cite{WTQ}, WIKISQL~\cite{WIKISQL}, and SQA~\cite{SQA}) across various approaches. 
In our experiment, we selected \texttt{google/tapas-large-finetuned}~\cite{tapas} and used four transformation methods: \textit{Column Adding}, \textit{Column Masking}, \textit{Column Order Shuffling}, and \textit{Row Order Shuffling}.

As shown in Fig. \ref{10}, \textbf{low difference between DEVAL and others.} DEVAL produces results largely consistent with existing robustness evaluation frameworks across different datasets and perturbation patterns
(with an average deviation of 3.1\% and a maximum deviation of 16.9\% in PromptBench, and an average deviation of 5.9\% and a maximum deviation of 18.3\% in RobuT).
The observed differences can be attributed to variations in modeling and implementation approaches under the same perturbation semantics. Meanwhile, we find \textbf{consistent trends across frameworks.}
The complexity of tasks in the RobuT framework leads to greater volatility in evaluation performance, but DEVAL still maintains a consistent trend with the RobuT framework under such conditions (with a correlation coefficient of 0.938), indicating a strong correlation. In summary, these results validate the effectiveness of DEVAL in the \textit{ID} setting, supporting its reliable extension to broader distributional evaluation beyond \textit{ID} case.

\subsection{Autonomous DR Generation}

To explore whether LLMs can autonomously generate high-quality DRs to alleviate the burden of manual rule construction for humans, we conducted a controlled evaluation. For each task covered in our dataset, we provided the LLM with detailed task descriptions, representative examples, and the formal concept of DR. The LLM was then prompted to generate 10 candidate DRs per task. These generated DRs were subsequently assessed by human annotators along five dimensions: descriptive rationality, formal correctness, implementability, use of domain knowledge, and whether the DR represents an identity transformation (\textit{ID}). The detailed annotation criteria are presented in Appendix D.

As shown in Fig. \ref{11}, although LLM-generated DRs often exhibit surface-level plausibility (e.g., achieving a 71.4\% rate in descriptive rationality), their formal correctness (38.6\%) and implementability (50.0\%) remain limited. Moreover, only 18.6\% of DRs demonstrate the use of meaningful domain knowledge, while 31.4\% fall into the trivial \textit{ID} category, highlighting a lack of depth in model-generated abstractions. Across task types such as logic and image-based reasoning, the quality of DRs further deteriorates, particularly in correctness and applicability. These findings indicate that, while LLMs may offer creative inspiration during early-stage DR formulation, the generated rules are not yet reliable for direct deployment without human validation. Therefore, for now, we recommend manual construction of DRs to ensure evaluation integrity. Enhancing LLMs' ability to autonomously synthesize abstract transformation rules remains a promising direction for future work.

\subsection{Knowledge-Telling vs. Fine-Tuning}
To compare whether explicitly telling LLMs about DR knowledge (via DP) or directly binding DR relations through SFT is more effective, we conducted a controlled comparative experiment using GPT-3.5-turbo as the base model. Specifically, we fine-tuned GPT-3.5-turbo on the entire derivation dataset, with 30\% of the data used for training and another 30\% held for testing.
Given that DR evaluation inherently involves input transformations and corresponding changes in output, we could not simply perform SFT using isolated “question–answer” pairs. To address this, we concatenated the two-round inputs and their corresponding outputs into a single “question–answer” record, allowing us to examine whether the model could acquire DR capabilities during the SFT process.

\begin{figure}[t]
\centering
  \includegraphics[width=0.476\textwidth]{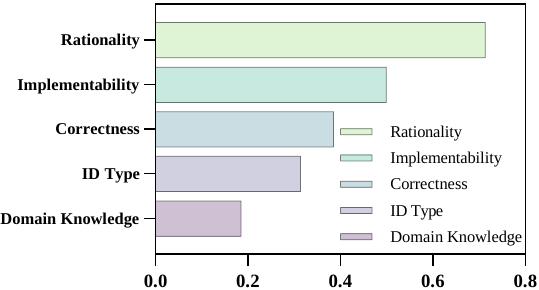}
\caption{Quality assessment of LLM‑generated DRs across five dimensions}\label{11}
     
\end{figure}

\begin{figure}[t]
\centering
  \includegraphics[width=0.476\textwidth]{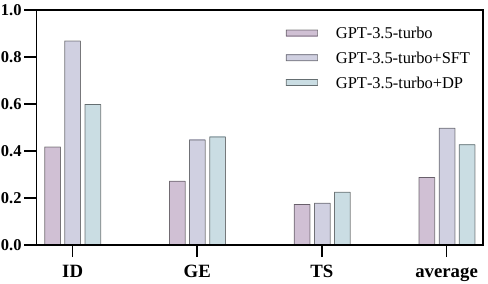}
\caption{Comparison of GPT‑3.5-turbo performance under DP and SFT.}\label{12}
     
\end{figure}

Fig. \ref{12} reveals nuanced differences in LLM behavior. SFT yields a substantial improvement on \textit{ID} transformations, raising performance from 41.9\% to 87.0\%, suggesting that SFT can enable LLMs to acquire a considerable degree of robustness. However, in tasks that require output adaptation, the fine-tuned GPT-3.5-turbo shows only marginal improvements and, in fact, underperforms compared to DP‑prompted GPT‑3.5-turbo, even in scenarios where fine-tuning is typically expected to outperform prompt engineering~\cite{Shin2023prompt_vs_finetuning}.
While fine-tuning achieves the highest overall average (49.9\%), its gains are almost entirely concentrated in the \textit{ID} subset. These findings indicate that \textbf{standard SFT, even when DR‑aware data is provided, does not substantially enhance LLMs' ability to understand abstraction-based reasoning behind DR.}
Therefore, relying solely on reference‑based evaluation and improvement is insufficient for abstract knowledge injection in LLMs. More symbolically grounded approaches are needed, which underscore the significance of our proposed DC.

\section{Limitation}
While the DEVAL framework provides a systematic approach to evaluating and improving DC, several limitations remain in its current form. These limitations span dataset coverage, attribution mechanisms, and quantifying the impact of task and DR difficulty. We discuss these challenges in detail below and outline potential directions for overcoming them in future work.

\textbf{Dataset coverage.}
Although the datasets used in DEVAL cover seven representative task types and include DRs of varying complexity (ID, GE, TS), they do not exhaustively capture the entire space of real-world scenarios. The evaluation cases are designed to be controlled and interpretable, which sometimes limits their coverage breadth. In future work, we aim to mitigate this limitation by using LLMs themselves to generate more diverse evaluation examples under human supervision. This could further expand the applicability and realism of the DEVAL framework.

\textbf{Attribution challenges.}
Our current method for error attribution relies heavily on LLM-generated CoT reasoning and LLM-based output labeling. While this approach provides insights into failure patterns, it lacks the reliability and credibility of a human-expert interpretability framework. At present, we are not aware of a more convincing and suitable alternative for reasoning attribution. Nevertheless, the current approach cannot fully reveal the dependency between CoT reasoning and the final conclusions. Future work may involve hybrid approaches that combine symbolic tracing, white-box analysis, or human annotation to improve both interpretability and attribution reliability.

\textbf{Task and DR Difficulty.}
Our evaluation results indicate that LLM performance varies significantly with both task difficulty and the complexity of derivation rules (DR). On the DR side, when transformation rules are simple (e.g., T = R) or only require the output unchanged (ID), LLMs tend to perform well. However, performance degrades notably for cases involving deeper semantic understanding or task-specific transformations. Similarly, the intrinsic difficulty of the task has a substantial impact on performance. Tasks involving multi-hop reasoning (e.g., logic and mathematics exhibit greater performance variance. These findings point to the need for modeling the topological relationship between task difficulty and DR types, which may enable a more systematic and hierarchical approach to DC development in future research.

\section{Related Work}
Much research on evaluating and improving performance on certain capabilities of LLM like robustness and counterfactual reasoning ability can be considered as special cases of the DC defined in this paper. 

\textbf{Robustness} is a typical case of DC that requires small changes of inputs resulting in identical or stable output. Zhu et al. proposed PromptRobust~\cite{PromptRobust} and PromptBench~\cite{PromptBench}, frameworks to dynamically evaluate robustness using four levels of adversarial attack. Wang, S. et al. introduced ReCode~\cite{ReCode}, noting that minor prompt modifications can lead to drastically different results. Wang, J et al.~\cite{Wang} evaluated ChatGPT's robustness under adversarial and out-of-distribution (OOD) conditions, examining output consistency with typos and distractions. Zhao, Y. et al. developed RobuT~\cite{RobuT}, testing conclusion consistency by interfering with table rows and columns. Stolfo, A. et al.~\cite{Stolfo} analyzed causal patterns in reasoning tasks by altering various parts of the question, focusing on sensitivity to interventions in mathematical problems. Chen, X. et al.~\cite{Chen} used linguistic transformations from TextFlint to show robustness performance across LLMs via the degradation rate. Li, C. et al. proposed R-BENCH~\cite{R-Bench}, a robustness datasets that target multimodal large models under real-world distortions.

\textbf{Counterfactual Reasoning} is another approach to reasoning, which evaluates if LLMs maintain performance under counterfactual conditions.
Miceli-Barone et al.~\cite{Miceli-Barone} found that even when identifiers are renamed with the same meaning in code generation tasks, LLMs still lack deep code understanding. Srivastava, S. et al.~\cite{Srivastava} evaluated the robustness in reasoning by altering input values, observing notable performance drops. Shapira, N.~\cite{Shapira} explored Nested Theory of Mind (N‑ToM) abilities in LLMs and found that they struggle with counterfactuals, relying on shallow heuristics rather than a robust theory of mind. Wu, Z. et al.~\cite{Wu} proposed an evaluation framework using task variants deviating from standard assumptions, showing that abstract reasoning of LLMs is not universally transferable. Lewis, M. et al.~\cite{Lewis} examined analogical reasoning and created counterfactual task variants, finding that GPT models performed poorly on these counterfactual challenges.

\section{Conclusion and Future Work}
In this paper, we define Derivation Capability (DC) as a novel property to evaluate the ability to recognize and maintain abstract rule consistency across different problem
transformations, and introduce a framework called DEVAL. DEVAL systematically assesses DC through three steps: Rule Formalization, Dataset Construction,
and Model Evaluation. It is applied to five mainstream LLMs and one Large Reasoning Model, revealing generally poor performance, with GPT-4o and Claude3.5 achieving average scores of 59.6\% and 52.5\%, respectively.
To address this, we propose Derivation Prompting, a novel prompt engineering approach that improves DC by an average of 15.2\%, highlighting its superiority over
five other commonly used methods. 

In future work, we propose to continue investigating more methods for improving DC, aiming to advance the alignment between LLMs and humans in understanding of abstract reasoning.

\bibliography{refs}
\onecolumn
\section*{Appendix}
\section*{A. Formalization of Derivation Relations in External Robustness Frameworks}
\label{A}
This section provides the formal definitions of the Derivation Relations (DRs) corresponding to the eight perturbations examined in Sec. 5.  
Each DR is meticulously crafted to align with the semantical nuances of the natural language descriptions of the respective perturbations and is implemented within the DEVAL framework.  
Among these, four DRs: \textit{CheckList}, \textit{StressTest}, \textit{TextBugger}, and \textit{DeepWordBug} are applied to natural language sentiment judgement tasks, including datasets \textit{SST-2}, \textit{QQP}, and \textit{MRPC}.  
The other four DRs: \textit{Column Adding}, \textit{Column Masking}, \textit{Column Order Shuffling}, and \textit{Row Order Shuffling} are implemented in table-based question-answering tasks, including datasets \textit{WTQ}, \textit{WikiSQL}, and \textit{SQA}.  
For each task, we provide its description, formal definition, and a representative instance. Additionally, for each DR, we present its description, domain-specific definition, and the associated implementation code
in both $T$ (data generation) and $R$ (result evaluation) sides to ensure reproducibility.

Note that in this section, we inspect the robustness property of the LLMs, so we provide a concrete definition of transformation $T$ for each task and define the transformation $R$ for every case as the identity relation, that is, $R=\{(y,y)\mid y\in Y\}$ where $Y$ is the output domain.
Also, the evaluation script is omitted in this section, as it is simply checking the equivalence.

\subsection*{1. Natural Language Semantical Classification}
\textbf{Task Description:} The task provides a set of natural language texts, requiring the LLM to determine the sentiment polarity of each text in \textit{SST-2}, and
to determine whether two texts are semantically equivalent in \textit{QQP} and \textit{MRPC}. The prompt in \textit{SST-2} is designed as follows: 
\texttt{\texttt{"As a sentiment classifier, determine whether the following text is "positive" or "negative". Please classify:\textbackslash n Question:\{content\} \textbackslash n  Answer:"}}

\noindent \textbf{Formal Definition:} Let $pr$ be the prompt above, $Q$ be the set of questions, where each question $q_i$ is a text in natural language.
Let $A = \{\texttt{positive}, \texttt{negative}\}$ be the set of answer options. We define the input domain $X$ and the output domain $Y$ as follows:
\begin{itemize}
  \item $X=\{(pr, q)\mid q \in Q\}$
  \item $Y=A$
\end{itemize}

\noindent \textbf{Representative Instance:}

\definecolor{orange}{RGB}{68,65,65}

\begin{center}
\begin{tcolorbox}[colback=gray!10,%gray background
  colframe=orange,% black frame colour
  width=18cm,% Use 5cm total width,
  arc=2mm, auto outer arc,
  title={Natural Language Semantical Classification},breakable,]
  \definecolor{red}{RGB}{89,38,57}
  \definecolor{green}{RGB}{32,69,85}
  \definecolor{yellow}{RGB}{91,90,37}

  \texttt{\textcolor{red}{\textbf{<Question>}}}

  % \texttt{for those moviegoers who complain that they don't make movies like they used to anymore}
  $\mathtt{...\ for\ those\ moviegoers\ who\ complain\ that\ they\ don't\ make\ movies\ like\ they\ used}$
  $\mathtt{to\ anymore}$

  \tcbline
  \texttt{\textcolor{green}{\textbf{<Answer>}}}

  % \texttt{negative}
  $\mathtt{negative}$
\end{tcolorbox}
\end{center}

\subsection*{\textbf{DR A.1.1: CheckList}}
\textbf{DR Description:} This DR will add 10 random characters as input perturbations to the end of each prompt.

\noindent \textbf{Formal Definition:} Let $Alphabet = \{a, b, ..., z, 0, 1, ..., 9\}$ be the set of random characters, and $CL = Alphabet^*$
%\{(n_1,n_2,...,n_{10})\mid n_i \in Alphabet, i \in \{1,2,...,10\} \}$ 
be the set of perturbation strings. Transformation $T$ is defined as follows:
% \\[1ex]
% \hspace*{5mm} $T = \{(x_1,x_2) \mid  \exists q\in Q, cl\in CL \cdot ( x_1 = (pr, q)\wedge x_2 = (pr+cl, q) ) \}$\\[1ex]
\begin{itemize}
  \item $T = \{(x_1,x_2) \mid  \exists q\in Q, cl\in CL \cdot ( x_1 = (pr, q)\wedge x_2 = (pr+cl, q) ) \}$
\end{itemize}
\textbf{Data Generation Script:}
\begin{lstlisting}
import random
def DR_A_1_1(prompt):
    ALPHABET = "abcdefghijklmnopqrstuvwxyz0123456789"
    CL = "".join([ALPHABET[int(random() * len(ALPHABET))] for _ in range(10)])
    return prompt+CL
\end{lstlisting}

\subsection*{\textbf{DR A.1.2: DeepWordBug}}

\textbf{DR Description:} This DR will add the letter ``s" after every letter ``a" and remove all occurrences of the letter ``t" in the prompt as a perturbation.

\noindent \textbf{Formal Definition:} Let $pr' = $\texttt{"Ass as senimen classifier, deermine if the following ex is "positive" or "negative". Please classsify:\textbackslash n Quesion:\{content\} \textbackslash n Asnswer:"} be the perturbation prompt. The transformation $T$ is defined as follows:
\begin{itemize}
  \item $T = \{(x_1,x_2) \mid \exists q \in Q \cdot (x_1 = (pr, q) \wedge x_2 = (pr', q))\}$
\end{itemize}
\textbf{Data Generation Script:}
\begin{lstlisting}
def DR_A_1_2_(prompt):
    prompt = prompt.replace("a", "as").replace("t", "")
    return prompt
\end{lstlisting}

\subsection*{2. Table-based Question-Answering}
\textbf{Task Description:} 
The task provides a data table and the corresponding questions related to the table. The LLM is required to retrieve the correct answer to the question by referring to the table.
In this task the prompt is embedded with table and questions to the LLM.

\noindent \textbf{Formal Definition:} Let $Table = \{1, 2, \ldots, m\} \times \{1, 2, \ldots, n\} \to \Sigma^*, m,n \in \mathbb{N}^+$ be the table to be referenced, and $Q$ be the set of questions. The notation
$(q,table)$ represents a pair consisting of a corresponding table and a question. We have task input $X$ and output $Y$ provided from the following definition:
\begin{itemize}
  \item $X =\{(q,table) \mid q\in Q, table \in Table\}$
  \item $Y =\{ans \mid \exists i \in \{1,2,...,m\}, j \in \{1,2,...,n\} \cdot ans = table(i,j)\}$
\end{itemize}

\noindent \textbf{Representative Instance:}
\begin{center}
  \begin{tcolorbox}[colback=gray!10,%gray background
    colframe=orange,% black frame colour
    width=18cm,% Use 5cm total width,
    arc=2mm, auto outer arc,
    title={Table-based Question-Answering},breakable,]
    \definecolor{red}{RGB}{89,38,57}
    \definecolor{green}{RGB}{32,69,85}
    \definecolor{yellow}{RGB}{91,90,37}
  
    \texttt{\textcolor{red}{\textbf{<Question>}}}
  
    % \texttt{Which nation won gold but did not win silver? table: \{header: [Rank, Nation, Gold, Silver, Bronze, Total], rows: [[1, Cuba, 4, 3, 2, 9], [2, Canada, 4, 2, 1, 7], [3, United States, 2, 0, 2, 4], [4, Mexico, 1, 1, 0, 2], [Total, Total, 12, 12, 12, 36]]\}}
    $\mathtt{Which\ nation\ won\ gold\ but\ did\ not\ win\ silver?}$\\
    $\mathtt{table: \{header: [Rank, Nation, Gold, Silver, Bronze, Total],\ rows: [[1, Cuba, 4, 3, 2,9],[2, Canada, 4, 2, 1, 7],}$\\
    $\mathtt{[3, United States, 2, 0, 2, 4], [4, Mexico, 1, 1, 0, 2], [Total, Total,12, 12, 12, 36]]\}}$

    \tcbline
    \texttt{\textcolor{green}{\textbf{<Answer>}}}
  
    % \texttt{United States}
    $\mathtt{United\ States}$
  \end{tcolorbox}
  \end{center}
\subsection*{\textbf{DR A.2.1: Column Adding}}

\textbf{DR Description:} This DR will add a new column ``Fare'' in the middle of the table as a perturbation, with the same values ``\$100'' for all rows.

\noindent \textbf{Formal Definition:} Let $CA= \{1, 2, \ldots, m\} \times \{\texttt{"}\mathtt{Fare}\texttt{"}\} \to \{\texttt{"}\mathtt{\$100}\texttt{"}\}$ be the perturbation to the original table. Transformation $T$ is defined as follows:
\begin{itemize}
  \item $T = \{(x_1,x_2) \mid  \exists q \in Q ,table \in Table \cdot ( x_1 = (q,table)\wedge  x_2 = (q,table \cup CA) ) \}$
\end{itemize}

\noindent \textbf{Data Generation Script:}

\begin{lstlisting}
import random
def DR_A_2_1(data):
    colomn = random.randint(len(data["table"]["header"]))
    header = data["table"]["header"]
    header = header[:colomn] + ["Fare"] + header[colomn:]
    data["table"]["header"] = header
    rows = data["table"]["rows"]
    for i in range(len(rows)):
        rows[i] = rows[i][:colomn] + ["$100"] + rows[i][colomn:]
    data["table"]["rows"] = rows
\end{lstlisting}

\subsection*{\textbf{DR A.2.2: Column Order Shuffling}}

\textbf{DR Description:} This DR reverses the order of all columns in the table as a perturbation.

\noindent \textbf{Formal Definition:} Let $table(a,b)$ be the value of the table in the $a$-th row and $b$-th column. Transformation $T$ is defined as follows:
\begin{itemize}
  \item $T = \{(x_1,x_2) \mid \exists q \in Q,table_1,table_2 \in Table \cdot ( x_1 = (q,table_1)\wedge x_2 = (q,table_2)$ $\wedge \forall i \in \{1,…,m\}, j \in \{1,…,n\} \cdot table_2(i,j) = table_1(i,n-j+1))\}$
\end{itemize}

\noindent \textbf{Data Generation Script:}

\begin{lstlisting}
def DR_A_2_2(data):
  header = data["table"]["header"]
  header = header[::-1]
  data["table"]["header"] = header
  rows = data["table"]["rows"]
  for i in range(len(rows)):
      rows[i] = rows[i][::-1]
  data["table"]["rows"] = rows
\end{lstlisting}

\section*{B. Formal Definitions of Derivation Relations in Tasks}
\label{B}
In this section, we illustrate the process of constructing the Derivation Capability (DC) datasets through four representative examples discussed in Sec. 3.  
These examples correspond to the \textit{choice} \textit{image}, \textit{logic}, and \textit{algorithm} tasks as mentioned in the main text, while the remaining other tasks are processed following an identical methodology. Performance in all tasks is given in Fig. \ref{Performance}
Note that examples representing robustness have been provided in detail in Appendix A. Due to their similar definitions and evaluation processes, robustness-related examples are not repeated in this section.

\subsection*{1. Common Sense Choice}
\label{B.1}
\textbf{Task Description:} The task provides a set of questions about common sense and facts, each question containing five candidate answers.
The LLM is required to select the option that it believes to be correct. The prompt is designed as follows:
\texttt{"Question:\{content\} \textbackslash n  Options:\{content\}}

\noindent \textbf{Task Dataset:} We sampled 200 cases from the xplainLLM dataset [Chen et al., 2024],
a QA dataset of 24,204 instances. Each instance includes a common sense question, five answer234
choices, the correct answer label, and an explanation for the prediction. The LLM is required to235
provide the correct answer based on the questions.

\noindent \textbf{Formal Definition:} Let $Q$ be the set of questions, $Ans = \{A, B, C, D, E\}$ be the set of answer options. For a question $q\in Q$, $Con(q)= Ans \rightarrow Strings$ is the family of functions from the option indexes to their contents.
%$Con=\{con_A, con_B, con_C, con_D, con_E\}$ be the content of each option.
We define the input domain $X$ and the output domain $Y$ as follows:

\begin{itemize}
  \item $X=\{(q,con)\mid q\in Q, con\in Con(q)\}$
  \item $Y=Ans$
\end{itemize}

\noindent \textbf{Representative Instance:}

\begin{center}
  \begin{tcolorbox}[colback=gray!10,%gray background
    colframe=orange,% black frame colour
    width=18cm,% Use 5cm total width,
    arc=2mm, auto outer arc,
    title={Common Sense Choice},breakable,]
    \definecolor{red}{RGB}{89,38,57}
    \definecolor{green}{RGB}{32,69,85}
    \definecolor{yellow}{RGB}{91,90,37}
  
    \texttt{\textcolor{red}{\textbf{<Question>}}}
  
    % \texttt{What is likely to have a better school cafeteria? A. high school, B. canteen, C. polytechnic, D. large room, E. all kinds of schools.}
    $\mathtt{What\ is\ likely\ to\ have\ a\ better\ school\ cafeteria?}$\\
    $\mathtt{A. high\ school,\ B. canteen,\ C. polytechnic,\ D. large\ room,\ E. all\ kinds\ of\ schools.}$

    \tcbline
    \texttt{\textcolor{green}{\textbf{<Answer>}}}
  
    % \texttt{C}
    $\mathtt{C}$
  \end{tcolorbox}
  \end{center}

\subsection*{\textbf{DR B.1.1: Option Shuffling}}
\label{B.1.1}
\textbf{DR Description:} This DR will shuffle the order of the answer options in the question while keeping the content of each option unchanged.

\noindent \textbf{Formal Definition:} The transformations $T$ and $R$ are defined as follows:

\begin{itemize}
  \item $T = \{(x_1,x_2) \mid \exists q \in Q ,con_1,con_2\in Con(q) \cdot(x_1 = (q,con_1)\wedge x_2 = (q,con_2) \wedge con_1(A) = con_2(E)\wedge con_1(B) = con_2(D)\wedge con_1(C) = con_2(C)\wedge con_1(D) = con_2(B)\wedge con_1(E) = con_2(A))\}$
  \item $R = \{(A,E), (B,D), (C,C), (D,B), (E,A)\}$
\end{itemize}

\noindent \textbf{Data Generation Script:}

\begin{lstlisting}
def DR_B_1_1(question, answer_dict)
    return (question, {k: answer_dict[v] for k, v in zip("ABCDE", "EDCBA")})
\end{lstlisting}

\noindent \textbf{Evaluation Script:}

\begin{lstlisting}
def DR_B_1_1_eval(ans1, ans2):
    return ord(ans1) + ord(ans2) == 2*ord('C')
\end{lstlisting}

\subsection*{\textbf{DR B.1.2: Explanation Addition}}
\label{B.1.2}
\textbf{DR Description:} In the task dataset, each question is also provided with a reasonable explanation for one of the answers that should not have been selected. This DR
adds the explanation to the description of the question, thereby changing the conditions of the question.

\noindent \textbf{Formal Definition:} For a question $q\in Q$, $Exp(q)= Ans \rightarrow Strings$ is the family of functions from the option indexes to their explanations.
Transformations $T$ and $R$ are defined as follows for all $ans \in Ans$:

\begin{itemize}
  \item $T = \{(x_1,x_2) \mid \exists q \in Q ,con \in Con(q),exp \in Exp(q) \cdot(x_1 = (q,con), x_2 = (q+exp(ans),con))\}$
  \item $R = \{(y_1,y_2) \mid y_2 = ans\}$
\end{itemize}

\noindent \textbf{Data Generation Script:}

\begin{lstlisting}
def DR_B_1_2(question, answer_dict, explanation)
    return (question+explanation["content"], answer_dict)
\end{lstlisting}

\noindent \textbf{Evaluation Script:}

\begin{lstlisting}
def DR_B_1_2_eval(ans1, ans2, explanation)
    return ans2 == explanation["option"]
\end{lstlisting}

\subsection*{2. Image Recognition}
\label{B.2}
\textbf{Task Description:} The task provides images of vehicles on a traffic scenario, with each vehicle facing left or right. The LLM is required to count the number of vehicles facing the left and the right in the image.
The prompt is simply the url of the tested image.

\noindent \textbf{Task Dataset:} We created a custom dataset of 200 instances. Each case contains an image of vehicles in a real-world road scene, positioned randomly with each vehicle facing either left or right, and vehicles do not overlap each other. The LLM is required to count the number of vehicles facing the left and the right in the image.

\noindent \textbf{Formal Definition:} Let $Pic \subseteq \{1, 2, \ldots, m\} \times \{1, 2, \ldots, n\} \to \{0, 1\}$ be the set of images to be recognized. We have task input $X$ and output $Y$ provided from the following definition:
\begin{itemize}
  \item $X=Pic$
  \item $Y=\{(L,R)\mid L,R \in \mathbb{N}\}$
\end{itemize}

\textbf{Representative Instance:}

\begin{center}
  \begin{tcolorbox}[colback=gray!10,%gray background
    colframe=orange,% black frame colour
    width=18cm,% Use 5cm total width,
    arc=2mm, auto outer arc,
    title={Image Recognition},breakable,]
    \definecolor{red}{RGB}{89,38,57}
    \definecolor{green}{RGB}{32,69,85}
    \definecolor{yellow}{RGB}{91,90,37}
  
    \texttt{\textcolor{red}{\textbf{<Question>}}}

    \includegraphics[width=1\textwidth]{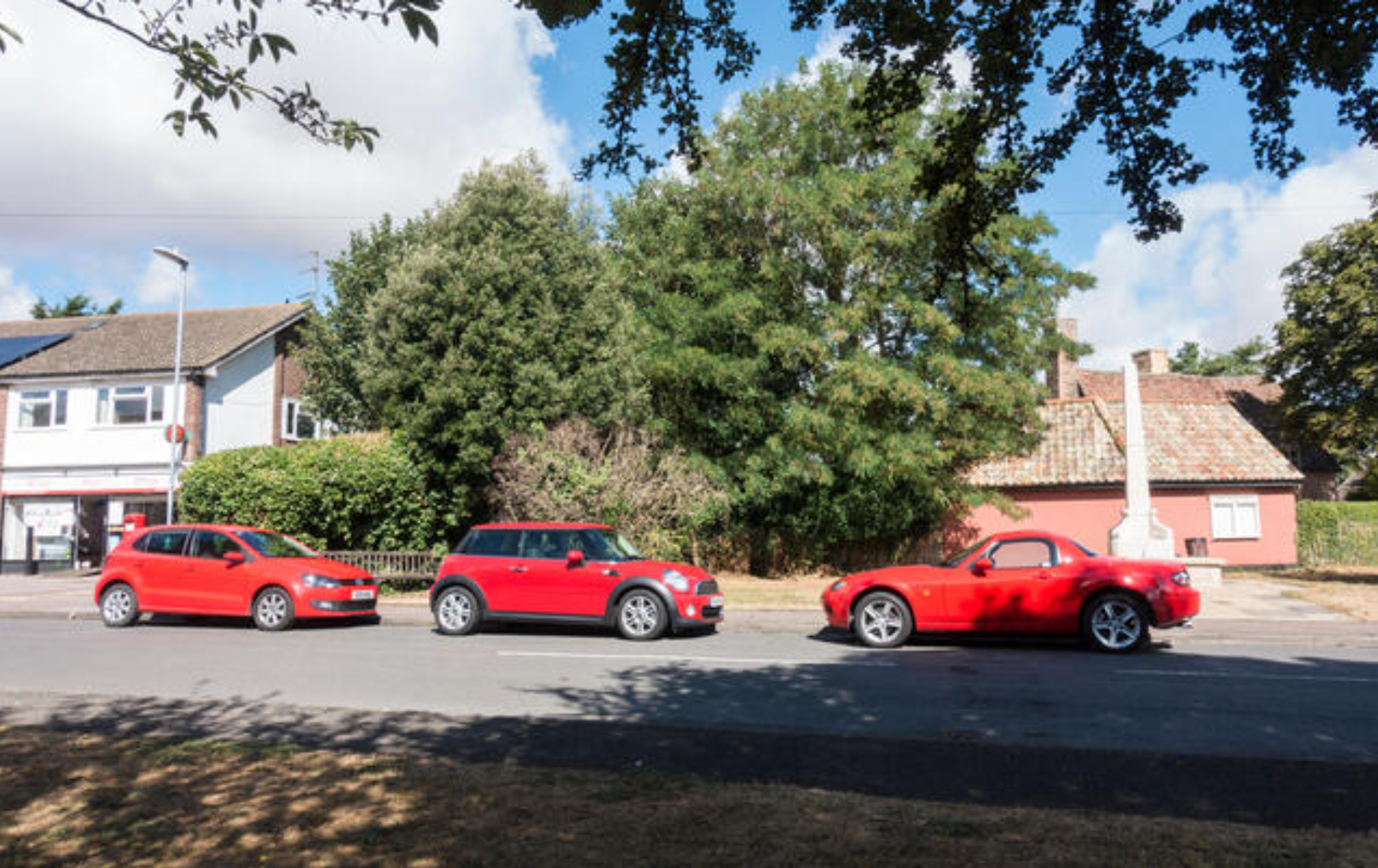}

    \tcbline
    \texttt{\textcolor{green}{\textbf{<Answer>}}}
  
    % \texttt{left:1 right:2}
    $\mathtt{left:1\quad right:2}$
  \end{tcolorbox}
  \end{center}

\subsection*{\textbf{DR B.2.1: Horizontal Flipping}}
\label{B.2.1}
\textbf{DR Description:} This DR will flip the image horizontally, changing the left-facing vehicles to right-facing vehicles and vice versa.

\noindent \textbf{Formal Definition:} Let $pic(a,b)$ be the pixel value of the image on the $a$ -th row and $b$ -th column. Transformations $T$ and $R$ are defined as follows:
\begin{itemize}
  \item $T = \{(x_1,x_2) \mid x_1\in X, x_2\in X, \forall i \in \{1,2,…,m\}, j \in \{1,2,…,n\} \cdot ( x_2(i,j) = x_1(i,n-j+1) )\}$
  \item $R = \{(y_1,y_2) \mid \exists l,r \in \mathbb{N} \cdot ( y_1 = (l,r)\wedge y_2 = (r,l))\}$
\end{itemize}

\noindent \textbf{Data Generation Script:}

\begin{lstlisting}
import cv2
def DR_B_2_1(img):
    return cv2.flip(img, 1)
\end{lstlisting}

\noindent \textbf{Evaluation Script:}

\begin{lstlisting}
def DR_B_2_1_eval(ans1, ans2):
    return ans1[0] == ans2[1] and ans1[1] == ans2[0]
\end{lstlisting}

\subsection*{\textbf{DR B.2.2: Vertical Concatenating}}
\label{B.2.2}
\textbf{DR Description:} This DR will concatenate two identical images vertically, doubling the number of vehicles in the image.

\noindent \textbf{Formal Definition:} Let $pic(a,b)$ be the pixel value of the image on the $a$ -th row and $b$ -th column. Transformations $T$ and $R$ are defined as follows:
\begin{itemize}
  \item $T = \{(x_1,x_2) \mid x_1 \in X, x_2\in X, \forall i \in \{1,…,m\}, j \in \{1,…,n\} \cdot (x_1(i,j) = x_2(i,j) \land x_2(i+m,j) = x_1(i,j))\}$
  \item $R = \{(y_1,y_2) \mid \exists l,r \in \mathbb{N} \cdot ( y_1 = (l,r), y_2 = (2*l,2*r) )\}$
\end{itemize}

\noindent \textbf{Data Generation Script:}

\begin{lstlisting}
import cv2
def DR_B_2_2(img):
    return cv2.vconcat([img, img])
\end{lstlisting}

\noindent \textbf{Evaluation Script:}

\begin{lstlisting}
def DR_B_2_2_eval(ans1, ans2):
    return ans1[0] == 2*ans2[0] and ans1[1] == 2*ans2[1]
\end{lstlisting}

\subsection*{3. Logical Reasoning}
\label{B.3}
\textbf{Task Description:} The task provides a set of reasoning problems, each consisting of a set of premises and a conclusion. The LLM is required to provide a proof chain to derive the conclusion from the premises.
The prompt is designed as follows:
\texttt{"Premises:\{content\}  Prove:\{content\}}

\noindent \textbf{Task Dataset:} We sampled 200 cases from the ProntoQA dataset, a first-order logic reasoning dataset with reasoning chains, where each case includes a set of natural language premises, a proof obligation, and a proof chain. The reasoning process involves multiple inference rules, such as AndElimination, AndIntroduction, OrElimination, and ProofByContradiction. We assigned numbers to each premise and required the LLM to output a list of used premise numbers in the correct inference order, along with a numbered proof chain as the label.

\noindent \textbf{Formal Definition:} Let $Prop$ be the set of propositions in terms of natural language. The input of the task is sequential in terms of a n+1 tuple $(p_1,..,p_n, q)$ with $p_1$ to $p_n$ being n premises and $q$ the conclusion.
The output is the minimal set of premise propositions from which the proof obligation can be derived, and the premise propositions are presented in the order of their introduction in the proof process.
Since we give each proposition a unique index, the output is a sequence of indexes of the propositions of the premise, denoted as $r_1, r_2, ..., r_m$, with $r_1$ to $r_{m-1}$ being the indexes of the propositions of the premise and $r_m$ the index of the conclusion.
%$Pre = \{[p_1,p_2,p_3,...,p_n] \mid i \in \{1,2,…,n\}, p_i \in \Sigma^*, n \in \mathbb{N} \}$ be the sets of question premises, where
%each premise $p_i$ is a natural language proposition. Let $C = \{c \in c_i \in \Sigma^* \}$ be the set of conclusions, where each conclusion $c$ is a natural language text.
%Let $<pre, c>$ be the pair of premises and conclusion which can be deduced from the premises.
%Let $Post = \{[q_1,q_2,q_3,...,q_m] \mid j \in \{1,2,…,m\}, q_j \in \Sigma^*, m \in \mathbb{N} \}$ be the prove chain to derive the conclusion from the premises.
We have task input $X$ and output $Y$ provided from the following definition:
\begin{itemize}
  \item $X=\{(p_1,p_2,...,p_n,q) \mid n\in \mathbb{N}, p_1\in Prop, ..., p_n\in Prop, q\in Prop\}$
  \item $Y=\{(r_1,r_2,...,r_m)\mid m \in \mathbb{N}, r_1\in \mathbb{N}, ..., r_m\in \mathbb{N},r_m=n+1\}$
\end{itemize}

\noindent \textbf{Representative Instance:}

\begin{center}
  \begin{tcolorbox}[colback=gray!10,%gray background
    colframe=orange,% black frame colour
    width=18cm,% Use 5cm total width,
    arc=2mm, auto outer arc,
    title={Logical Reasoning},breakable,]
    \definecolor{red}{RGB}{89,38,57}
    \definecolor{green}{RGB}{32,69,85}
    \definecolor{yellow}{RGB}{91,90,37}
  
    \texttt{\textcolor{red}{\textbf{<Question>}}}
  
    % \texttt{(1) Sterpuses are numpuses and dumpuses. (2) Every rompus is a sterpus and a zumpus. (3) Each numpus is a gorpus and a tumpus. (4) Rex is a sterpus and a numpus. Prove: (5) Rex is a sterpus.}
    $\mathtt{Premises:\ (1) Sterpuses\ are\ numpuses\ and\ dumpuses.\ (2) Every\ rompus\ is\ a\ sterpus\ and\ a\ zumpus.\ (3) Each}$\\
    $\mathtt{numpus\ is\ a\ gorpus\ and\ a\ tumpus.\ (4) Rex\ is\ a\ sterpus\ and\ a\ numpus.}$\\ 
    $\mathtt{Prove:\ (5) Rex\ is\ a\ sterpus.}$

    \tcbline
    \texttt{\textcolor{green}{\textbf{<Answer>}}}
  
    % \texttt{[(4), (5)]}
    $\mathtt{(4),\ (5)}$
  \end{tcolorbox}
  \end{center}

\subsection*{\textbf{DR B.3.1: Order Reassigning}}
\label{B.3.1}
\textbf{DR Description:} This DR will reverse the ordering of the premises in the reasoning problem.

\noindent \textbf{Formal Definition:} The transformations $T$ and $R$ are defined as follows:
\begin{itemize}
  \item $T = \{(x_1,x_2) \mid \exists p_1 , p_2 , ... , p_n , q \in Prop \cdot ( x_1 = (p_1 , p_2 , ... , p_n , q), x_2 = (p_n , p_{n-1} , ... , p_1 , q))\}$
  \item $R = \{(y_1,y_2) \mid \exists r_1 , ... , r_{m-1} \in \mathbb{N}^+ \cdot ( y_1 = (r_1 , ... , r_{m-1},n+1)\wedge y_2 = (n+1-r_{m-1} , ... , n+1-r_1, n+1))\}$
\end{itemize}

\noindent \textbf{Data Generation Script:}

\begin{lstlisting}
def DR_B_3_1(data):
    pre = data[0].split(".")
    pre.reverse()
    for i in range(0, len(pre)):
        pre[i] = pre[i].split(") ")[1]
        pre[i] = f"({i+1}) {pre[i].strip()}. "
    pre = "".join(pre).strip()
\end{lstlisting}

\noindent \textbf{Evaluation Script:}

\begin{lstlisting}
def DR_B_3_1_eval(ans1, ans2):
    return list(map(int, ans1)) == list(map(lambda x:len(pre)-int(x), ans2))
\end{lstlisting}

\subsection*{\textbf{DR B.3.2: Corollary}}
\label{B.3.2}
\textbf{DR Description:} This DR will add a new conclusion which can be derived from the original premises and a new premise saying that the original conclusion implies a new conclusion.

\noindent \textbf{Formal Definition:} Let $nq \in Prop$ be the new conclusion, $np \in Prop$ be the following premise: \texttt{"If $q$, then $nq$."}. Transformations $T$ and $R$ are defined as follows:
\begin{itemize}
  \item $T = \{(x_1,x_2) \mid \exists p_1 , p_2, ... , p_n , q,nq \in Prop \cdot (x_1 = (p_1 ,p_2, ... , p_n , q), x_2 = (np,p_1 , ... , p_n, nq))\}$
  \item $R = \{(y_1,y_2) \mid \exists r_1, ... , r_{m-1} \in \mathbb{N}^+ \cdot ( y_1 = (r_1 , ... , r_{m-1}, n+1), y_2 = (r_1 +1... , r_{m-1}+1,1, n+2))\}$
\end{itemize}

\noindent \textbf{Data Generation Script:}

\begin{lstlisting}
def DR_B_3_2(data):
    post = data[1]
    pre = data[0].split(".")
    post = f"Prove: ({len(pre)+2}) Taylor is a visionary."
    for i in range(0, len(pre)):
        pre[i] = pre[i].split(") ")[1]
        pre[i] = f"({i+2}) {pre[i].strip()}. "
    pre = "".join(pre).strip()
    pre = f"(1) if {data[1].split(') ')[1][:-1]} then Taylor is a visionary. {pre}"
\end{lstlisting}

\noindent \textbf{Evaluation Script:} 

\begin{lstlisting}
def DR_B_3_2_eval(ans1, ans2):
    return ans2[-2] == "(1)" and 
        list(map(lambda x:int(x)+1, ans1)) == list(map(int, ans2)).pop(-2)
\end{lstlisting}

\subsection*{4. Algorithm Simulation}
\label{B.4}
\textbf{Task Description:} The task provides undirected graphs with a set of nodes and edges. The LLM is required to find the path between two given nodes.
The prompt is designed as follows:
\texttt{"The undirected graph:\{content\} \textbackslash n The start node:\{content\} The end node:\{content\}"}

\noindent \textbf{Task Dataset:} We sampled 200 cases from the AQA-Bench dataset, an interactive benchmark designed to evaluate the sequential reasoning abilities of LLMs in algorithmic environments. One of its tasks involves the LLM completing path-search problems in an undirected graph, which includes a set of graph nodes, edges, and information about the start and end points. The LLM is tasked with finding the path from the start to the end point.

\noindent \textbf{Formal Definition:} Let $UG$ be the set of problem graphs with each graph having nodes $V = \{v_1, v_2, \ldots, v_n\}$ and edges $E \subseteq  V \times V$
and $\langle v_1,v_2,...,v_k \rangle $ be the path between $v_1$ and $v_k$.
We have task input $X$ and output $Y$ provided from the following definition:
\begin{itemize}
  \item $X = \{(graph,start,end) \mid graph \in UG ,start, end \in V\} $
  \item $Y = \{\langle v_1,v_2,...,v_k \rangle \mid v_1,v_2,...,v_k \in V , k \in \mathbb{N}^+, \forall i \in \{1,2,...,k-1\} \cdot (v_i,v_{i+1}) \in E\}$
\end{itemize}

\noindent \textbf{Representative Instance:}

\begin{center}
  \begin{tcolorbox}[colback=gray!10,%gray background
    colframe=orange,% black frame colour
    width=18cm,% Use 5cm total width,
    arc=2mm, auto outer arc,
    title={Algorithm Simulation},breakable,]
    \definecolor{red}{RGB}{89,38,57}
    \definecolor{green}{RGB}{32,69,85}
    \definecolor{yellow}{RGB}{91,90,37}
  
    \texttt{\textcolor{red}{\textbf{<Question>}}}
  
    % \texttt{tree: \{nodes: [0, 1, 2, 3, 4, 5, 6, 7], edges: [[0, 2], [1, 3], [2, 6], [2, 4], [3, 5], [3, 6], [4, 7]]\} start: 2 target: 5}
    $\mathtt{The\ undirected\ graph:\ \{nodes: [0, 1, 2, 3, 4, 5, 6, 7],\ edges: [[0, 2], [1, 3], [2, 6], [2, 4],[3, 5], [3, 6], [4, 7]]\}}$
    $\mathtt{The\ start\ node:\ 2\ The\ end\ node:\ 5}$
    \tcbline
    \texttt{\textcolor{green}{\textbf{<Answer>}}}
  
    % \texttt{2->6->3->5}
    $\mathtt{2\rightarrow 6\rightarrow 3\rightarrow 5}$
  \end{tcolorbox}
  \end{center}

\subsection*{\textbf{DR B.4.1: Point Swapping}}
\label{B.4.1}
\textbf{DR Description:} This DR swaps the start and end nodes of the problem.

\noindent \textbf{Formal Definition:} The transformations $T$ and $R$ are defined as follows:
\begin{itemize}
  \item $T = \{(x_1,x_2)  \mid \exists graph \in UG,v_1,v_2 \in V \cdot ( x_1 = (graph,v_1,v_2),  x_2 = (graph,v_2,v_1))\}$
  \item $R = \{(y_1,y_2) \mid y_1,y_2\in Y, \exists v_1,...,v_k \in V \cdot ( y_1 = \langle v_1,...,v_k \rangle, y_2 = \langle v_k,...,v_1 \rangle)\}$
\end{itemize}

\noindent \textbf{Data Generation Script:}

\begin{lstlisting}
def DR_B_4_1(graph, start, end):
    return (graph, end, start)
\end{lstlisting}

\noindent \textbf{Evaluation Script:}

\begin{lstlisting}
def DR_B_4_1_eval(ans1, ans2):
    return ans1.split("->") == ans2.split("->")[::-1]
\end{lstlisting}

\subsection*{\textbf{DR B.4.2: Endpoint Changing}}
\label{B.4.2}
\textbf{DR Description:} This DR changes the endpoint of the problem to the previous point in the path.

\noindent \textbf{Formal Definition:} The transformations $T$ and $R$ are defined as follows:
\begin{itemize}
  \item $T = \{(x_1,x_2) \mid \exists graph \in UG, v_1,...,v_k \in V \cdot ( k>1\wedge 
  x_1 = (graph,v_1,v_k)\wedge x_2 = (graph,v_1,v_{k-1})$ $\wedge \forall i \in \{1,2,…,k-1\} \cdot (v_i,v_{i+1}) \in E)\}$
  \item $R = \{(y_1,y_2) \mid y_1\in Y, y_2\in Y, \exists v_1,...,v_k \in V \cdot ( y_1 = \langle v_1,...,v_k \rangle, y_2 = \langle v_1,...,v_{k-1} \rangle)\}$
\end{itemize}

\noindent \textbf{Data Generation Script:}

\begin{lstlisting}
def DR_B_4_2(graph, start, end):
    return (graph, start, predecessor(graph, start, end))
\end{lstlisting}

\noindent \textbf{Evaluation Script:}

\begin{lstlisting}
  def DR_B_4_2_eval(ans1, ans2):
      return ans1.split("->") == ans2.split("->")[:-1]
\end{lstlisting}
\begin{figure*}
    \centering  
    \includegraphics[width=0.99\textwidth]{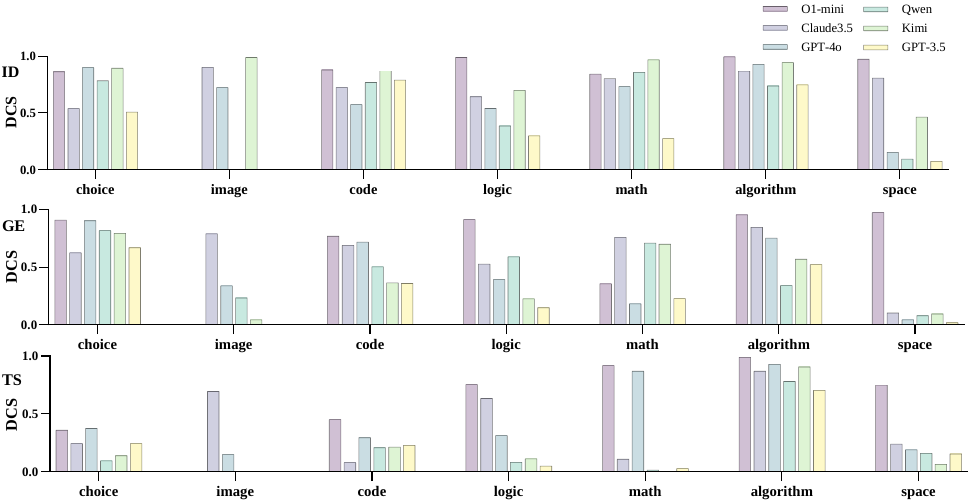}
    \caption{Derivation Capability performance of six LLMs in seven task scenarios.}\label{Performance}
    \vspace{-0.5cm}
  \end{figure*}

\section*{C. Prompt Engineering Methods Settings}
\label{C}
This section presents the PE settings for each method selected in Sec. 4.
Since all tasks and DRs follow a consistent pattern across other PE methods, we provide a unified description of the implementation for each method.

\textbf{Derivation Prompting (DP)}: We involved the prompt ``For the second question, you need to: (1) first explain what change has occurred in the input compared to the first question; (2) explain what kind of change in output should result from this change in input in the context of this task; (3) apply this output change to your original answer to the first question in order to solve the second question." into all examples, promoting LLMs to internalize and apply DR rules systematically.

\textbf{Chain-of-Thought (CoT)}: We incorporated the prompt ``For each problem, you need to provide your thought process." into all examples, encouraging LLMs to develop
a step-by-step reasoning approach, thereby improving their consistency in DC.

\textbf{Step-Back (SB)}: We added the prompt ``For each problem, you need to first explain the fundamental principles involved in solving it." to all examples,
guiding LLMs to explain the rule embedded in the questions to enhance their understanding.

\textbf{One-Shot (OS)}: We integrated a pair of original and transformed question-answer pairs from the dataset into the prompt, emphasizing the relation
within the data to deepen the understanding of DR.

\textbf{Few-Shot (FS)}: We included five pairs of original
and transformed question-answer pairs in the prompt to achieve better performance.

\textbf{Analogy (AN)}: We united the prompt ``For each problem, you need to first present three related but not identical problems, along with a description of each problem and its solution."
to all examples, activating the prior training knowledge to enhance DC.

\section*{D. Autonomous DR Generation}
In this section, we illustrate the ability of LLMs to generate DRs automatically in Sec. 5, intending to assess whether they can complete the DR construction process to reduce human involvement in the Rule Formalization stage of DEVAL. We use GPT-4o as the experimental model and perform DR generation for each task in our dataset. For each task, the GPT-4o is provided with the task description, representative question–answer examples, and the concept of DR itself (directly using the formal definition from Sec. 2, along with illustrative examples from the main text). Based on this information, the model is asked to generate candidate DRs, including both natural language descriptions and corresponding formal (T, R) definitions.

For each task, we generate 10 candidate DRs and conduct manual evaluation. The evaluation covers five dimensions:

\textbf{(A.)} Rationality of the DR description, i.e., whether the results' description is reasonable and can be considered a valid DR;

\textbf{(B.)} Formal correctness of the DR, i.e., whether the provided (T, R) satisfies the formal definition;

\textbf{(C.)} Whether it belongs to the ID type, meaning the output remains unchanged; 

\textbf{(D.)} Whether domain-specific knowledge is used in constructing the DR; and

\textbf{(E.)} Implementability, i.e., whether the DR can be practically executed.

We present all the generated DR summaries and our manual evaluation annotations in Table I.

\begin{table}[H]
\caption{Summary of autonomously generated DRs and corresponding manual annotations across all seven tasks.}
\centering
\label{tab:choice_dr}
\begin{tabularx}{\textwidth}{c l l X}
\toprule
\textbf{Task} & \textbf{Index} & \textbf{Annotation} & \textbf{DR Description} \\
\midrule
\multirow{10}{*}{Choice} 
 & \textbf{DR1} & A,B,E & Option Permutation – Shuffling the order of answer options should trigger corresponding label updates. \\
\cmidrule(l){2-4}
 & \textbf{DR2} & A,B,C,E & Semantically Equivalent Rewrite – Replace phrases in the question with synonyms; the answer remains unchanged. \\
\cmidrule(l){2-4}
 & \textbf{DR3} & -- & Double Negation Transformation – Introduce double negation in the question, flipping the answer where applicable. \\
\cmidrule(l){2-4}
 & \textbf{DR4} & A,C,E & Distractor Injection – Add an irrelevant option; the correct answer remains unchanged. \\
\cmidrule(l){2-4}
 & \textbf{DR5} & A,C & Question Clause Reordering – Reorder clauses in the question without changing semantics; the answer remains the same. \\
\cmidrule(l){2-4}
 & \textbf{DR6} & A,B,D,E & Negated Question Inversion – Add negation to the question so that the correct answer flips to the opposite option. \\
\cmidrule(l){2-4}
 & \textbf{DR7} & A,C & Option Content Simplification – Shorten long option texts to keywords; the answer remains unchanged. \\
\cmidrule(l){2-4}
 & \textbf{DR8} & A,B,C,E & Label Perturbation – Change labels (e.g., A–E → 1–5) without changing content; the mapping of the answer follows. \\
\cmidrule(l){2-4}
 & \textbf{DR9} & D & Single → Multiple with Single Answer – Change the question to multi-select format but only one option is correct. \\
\cmidrule(l){2-4}
 & \textbf{DR10} & A,B,C,E & Instructional Noise Injection – Add irrelevant instructions (e.g., “Read carefully”); the answer remains unchanged. \\
\midrule
\bottomrule
\end{tabularx}
\end{table}

\begin{table}[H]
\centering
\label{tab:image_dr}
\begin{tabularx}{\textwidth}{c l l X}
\toprule
\textbf{Task} & \textbf{Index} & \textbf{Annotation} & \textbf{DR Description} \\
\midrule
\multirow{10}{*}{Image} 
 & \textbf{DR1} & A,B,E & Flip the image horizontally; left-facing and right-facing vehicle counts should swap. \\
\cmidrule(l){2-4}
 & \textbf{DR2} & A,B,E & Vertically concatenate the same image; vehicle counts double while directions remain unchanged. \\
\cmidrule(l){2-4}
 & \textbf{DR3} & B & Crop the left half of the image; right-facing vehicles decrease or remain the same. \\
\cmidrule(l){2-4}
 & \textbf{DR4} & A,E & Add one extra right-facing vehicle; right-facing count increases by 1. \\
\cmidrule(l){2-4}
 & \textbf{DR5} & A,C,E & Apply blur without altering structure; output should remain unchanged. \\
\cmidrule(l){2-4}
 & \textbf{DR6} & A,C & Add noise without occluding vehicles; output should remain unchanged. \\
\cmidrule(l){2-4}
 & \textbf{DR7} & A,C,E & Increase image brightness; vehicle orientation recognition should remain unchanged. \\
\cmidrule(l){2-4}
 & \textbf{DR8} & A & Replace one left-facing vehicle with a right-facing vehicle; left decreases by 1 and right increases by 1. \\
\cmidrule(l){2-4}
 & \textbf{DR9} & -- & Block the center of the image; output becomes unpredictable (non-deterministic DR). \\
\cmidrule(l){2-4}
 & \textbf{DR10} & A,B,E & Horizontally concatenate the image twice; total vehicle count doubles. \\
\midrule
\midrule
\multirow{10}{*}{Code} 
 & \textbf{DR1} & A,E & Increment Analysis – Analyze incremental operations (e.g., k += 1) to derive related loop invariants. \\
\cmidrule(l){2-4}
 & \textbf{DR2} & -- & Assignment Derivation – Derive loop invariants from variable assignments (e.g., sum\_value += arr[k]). \\
\cmidrule(l){2-4}
 & \textbf{DR3} & A,B,E & Termination Condition Derivation – Derive loop boundary invariants from termination conditions (e.g., k < len(arr)). \\
\cmidrule(l){2-4}
 & \textbf{DR4} & A,B,E & Loop Boundary Analysis – Analyze array index changes in the loop to derive last accessed element invariants. \\
\cmidrule(l){2-4}
 & \textbf{DR5} & -- & Iteration Pattern Derivation – Identify iteration patterns (e.g., k += 1) to deduce loop invariants. \\
\cmidrule(l){2-4}
 & \textbf{DR6} & A,E & Array Size Influence – Derive invariants related to array size (e.g., k = len(arr) at termination). \\
\cmidrule(l){2-4}
 & \textbf{DR7} & -- & Accumulating Variable Change – Deduce how accumulation variables (e.g., sum\_value) evolve as invariants. \\
\cmidrule(l){2-4}
 & \textbf{DR8} & A,B,D,E & Control Flow Change – Derive invariants under modified control flow (e.g., break when k == len(arr)). \\
\cmidrule(l){2-4}
 & \textbf{DR9} & A,C,E & Comparison Operation Derivation – Derive invariants from comparison operations like k < len(arr). \\
\cmidrule(l){2-4}
 & \textbf{DR10} & -- & State Change Derivation – Deduce invariants reflecting state updates in the loop (e.g., sum\_value changes). \\
\midrule
\midrule
\multirow{10}{*}{Space} 
 & \textbf{DR1} & A,B,E & Direction Reversal – Change clockwise movement to counterclockwise (or vice versa); move to the opposite position with same steps. \\
\cmidrule(l){2-4}
 & \textbf{DR2} & B,E & Step Increase – If the step count increases by k, the final position should shift forward by k steps. \\
\cmidrule(l){2-4}
 & \textbf{DR3} & A,B,C,D,E & Modular Closure – All step changes follow modulo 20; position changes by steps mod 20. \\
\cmidrule(l){2-4}
 & \textbf{DR4} & A,C,E & Direction Consistency Merge – Consecutive clockwise moves can be merged into one with the total steps. \\
\cmidrule(l){2-4}
 & \textbf{DR5} & A,B,E & Start Shift Symmetry – Shifting the starting point by m causes the endpoint to shift by the same m. \\
\cmidrule(l){2-4}
 & \textbf{DR6} & C,D & Boundary Wrap-Around – When total steps exceed 20, wrap around according to modulo rules. \\
\cmidrule(l){2-4}
 & \textbf{DR7} & A,C & Direction Mixing – Alternate clockwise and counterclockwise steps reduce to net movement. \\
\cmidrule(l){2-4}
 & \textbf{DR8} & A,C & Unique Mapping of Positions – Sequences with the same net displacement map to the same final position. \\
\cmidrule(l){2-4}
 & \textbf{DR9} & A,B,E & Relative Position Invariance – Starting at the previous endpoint and performing the reverse sequence returns to the original start. \\
\cmidrule(l){2-4}
 & \textbf{DR10} & -- & No Movement – If the step count is zero, the final position remains unchanged. \\
\midrule
\midrule
\multirow{10}{*}{Math} 
 & \textbf{DR1} & A,B,E & Basic Polynomial Integration – Integrate a simple polynomial and evaluate at a specific value. \\
\cmidrule(l){2-4}
 & \textbf{DR2} & -- & Exponential Function Integration – Integrate an exponential function and apply transformation. \\
\cmidrule(l){2-4}
 & \textbf{DR3} & A,B & Input Shift – Apply a shift to the input variable and observe the integral’s behavior. \\
\cmidrule(l){2-4}
 & \textbf{DR4} & -- & Polynomial Integration – Integrate a polynomial and substitute x = 4. \\
\cmidrule(l){2-4}
 & \textbf{DR5} & A,B,D,E & Constant Replacement – Change the constant term and observe its effect on the integral. \\
\cmidrule(l){2-4}
 & \textbf{DR6} & A,D & Exponent Change – Modify the exponent of the variable and compute the integral. \\
\cmidrule(l){2-4}
 & \textbf{DR7} & A,B,D,E & Linear Transformation of Input – Apply linear scaling to input and observe corresponding changes. \\
\cmidrule(l){2-4}
 & \textbf{DR8} & A & Composite Function Integration – Integrate a composite function and check the result. \\
\cmidrule(l){2-4}
 & \textbf{DR9} & A,C,D,E & Reciprocal Function Integration – Compute the integral of a reciprocal function under transformation. \\
\cmidrule(l){2-4}
 & \textbf{DR10} & A,B,D,E & Trigonometric Function Integration – Integrate a trigonometric function with input transformation. \\
\midrule
\bottomrule
\end{tabularx}
\end{table}

\begin{table}[H]
\centering
\label{tab:math_dr}
\begin{tabularx}{\textwidth}{c l l X}
\toprule
\textbf{Task} & \textbf{Index} & \textbf{Annotation} & \textbf{DR Description} \\
\midrule
\multirow{10}{*}{Algorithm} 
 & \textbf{DR1} & A,B,E & Point Swapping – Swap the start and end nodes; the output path should be reversed. \\
\cmidrule(l){2-4}
 & \textbf{DR2} & A & Graph Structure Change – Reverse the direction of one edge; the shortest path updates accordingly. \\
\cmidrule(l){2-4}
 & \textbf{DR3} & -- & Node Order Shuffling – Shuffle the node order; the shortest path follows the new graph structure. \\
\cmidrule(l){2-4}
 & \textbf{DR4} & A & Edge Weight Change – Update edge weights; the output should reflect the new shortest path. \\
\cmidrule(l){2-4}
 & \textbf{DR5} & A,B,D,E & Path Reversal – Reverse the sequence of the shortest path to obtain a new path. \\
\cmidrule(l){2-4}
 & \textbf{DR6} & -- & Edge Deletion – Remove an edge; output should be the shortest path in the updated graph. \\
\cmidrule(l){2-4}
 & \textbf{DR7} & A & Node Addition – Add a new node and its edges; output should reflect the updated shortest path. \\
\cmidrule(l){2-4}
 & \textbf{DR8} & A,B,D,E & Target Change – Change the end node; the shortest path should update accordingly. \\
\cmidrule(l){2-4}
 & \textbf{DR9} & A,B,D,E & Start Change – Change the start node; the shortest path should update accordingly. \\
\cmidrule(l){2-4}
 & \textbf{DR10} & -- & Edge Replacement – Replace an edge; output should be the new shortest path. \\
\midrule
\midrule
\multirow{10}{*}{Logic} 
 & \textbf{DR1} & A,C,E & Order Reversing – Reverse the order of premises to verify the stability of the conclusion. \\
\cmidrule(l){2-4}
 & \textbf{DR2} & A,B,D,E & Premise Removal – Remove a premise to test whether the conclusion derivation still holds. \\
\cmidrule(l){2-4}
 & \textbf{DR3} & A,C,E & Premise Shuffling – Shuffle the order of premises to check the robustness of conclusion derivation. \\
\cmidrule(l){2-4}
 & \textbf{DR4} & -- & Conclusion Modification – Modify the conclusion to test dependency on premises. \\
\cmidrule(l){2-4}
 & \textbf{DR5} & -- & Premise Repetition – Repeat certain premises to observe the effect on conclusion derivation. \\
\cmidrule(l){2-4}
 & \textbf{DR6} & C & Premise Replacement – Replace one premise to verify whether the conclusion is still valid. \\
\cmidrule(l){2-4}
 & \textbf{DR7} & A,D & Inference Rule Application – Apply a specific inference rule to generate the new conclusion. \\
\cmidrule(l){2-4}
 & \textbf{DR8} & A,C & Conclusion Consistency – Check whether the conclusion remains consistent under premise changes. \\
\cmidrule(l){2-4}
 & \textbf{DR9} & A,C & Premise Logical Transformation – Transform the logical structure of a premise to test derivation stability. \\
\cmidrule(l){2-4}
 & \textbf{DR10} & -- & Premise Addition – Add a new premise to verify how the conclusion changes. \\
\midrule
\bottomrule
\end{tabularx}
\end{table}

\end{document}